\documentclass[review,sort&compress]{elsarticle}

\usepackage[caption=false,font=footnotesize]{subfig}

\usepackage{amsmath}
\usepackage{amssymb}
\usepackage{algorithmic}
\usepackage{array}
\usepackage{dblfloatfix}
\usepackage{balance}

\usepackage{url}
\usepackage{lineno,hyperref}
\modulolinenumbers[5]

\usepackage{graphicx}
\graphicspath{{./figures/jpeg/}}
\DeclareGraphicsExtensions{.pdf,.jpeg,.png,.jpg}

\usepackage{xcolor}
\usepackage{color,soul}

\newcommand{\etal}{\textit{et al.}}
\newcommand{\ie}{\textit{i}.\textit{e}.}
\newcommand{\eg}{\textit{e}.\textit{g}.}
\newcommand{\wrt}{\textit{w}.\textit{r}.\textit{t}.}

\usepackage{times}

\usepackage{afterpage}
\usepackage{multirow}

\begin{document}

\begin{frontmatter}
\title{Deep Feature Augmentation for Occluded Image Classification}

\author[tongjiaddress]{Feng~Cen\corref{correspondingauthor}}
\cortext[correspondingauthor]{Corresponding author}
\ead{feng.cen@tongji.edu.cn}
\author[tongjiaddress]{Xiaoyu~Zhao}
\author[tongjiaddress]{Wuzhuang~Li}
\author[guanghuiaddress]{Guanghui~Wang}
\ead{ghwang@ku.edu}

\address[tongjiaddress]{The Department of Control Science \& Engineering, College of Electronics and Information Engineering, Tongji University, Shanghai 201804, China}
\address[guanghuiaddress]{Department of Computer Science, Ryerson University,
Toronto, ON, Canada M5B 2K3}

\begin{abstract}
	Due to the difficulty in acquiring massive task-specific occluded images, the classification of occluded images with deep convolutional neural networks (CNNs) remains highly challenging.
	To alleviate the dependency on large-scale occluded image datasets, we propose a novel approach to improve the classification accuracy of occluded images by fine-tuning the pre-trained models with a set of augmented deep feature vectors (DFVs). The set of augmented DFVs is composed of original DFVs and pseudo-DFVs. The pseudo-DFVs are generated by randomly adding difference vectors (DVs), extracted from a small set of clean and occluded image pairs, to the real DFVs. In the fine-tuning, the back-propagation is conducted on the DFV data flow to update the network parameters.
	The experiments on various datasets and network structures show that the deep feature augmentation significantly improves the classification accuracy of occluded images without a noticeable influence on the performance of clean images.
	Specifically, on the ILSVRC2012 dataset with synthetic occluded images, the proposed approach achieves 11.21\% and 9.14\% average increases in classification accuracy for the ResNet50 networks fine-tuned on the occlusion-exclusive and occlusion-inclusive training sets, respectively.
\end{abstract}


\end{frontmatter}

\section{Introduction}\label{sec:intro} 
\vspace{-6pt}
As an intractable intra-class variation for image classification, partial occlusion is ubiquitous in real-world images. 
Many research attempts have been made to address this problem.
Some researchers explore the information of the occluded object, the occludee, to design occlusion-robust representation of the images~\cite{liu1990partial, zhang2003object,huang1997object,ohba1997detectability}; while others focus on the intervening object, the occluder, to pursue an accurate model of the occluders so as to alleviate the influence of the occlusion~\cite{yang2013gabor,ou2014robust,yu2017discriminative,zhou2009face,jia2012robust,li2013structured,yang2017nuclear}.
Unfortunately, for classical approaches without employing the deep networks, promising results were only achieved in specific applications, such as face recognition on small datasets under constrained environments like indoor and front-view \cite{deng2012extended, yang2017nuclear, wu2018occluded, yu2017discriminative}.

Although recent deep convolutional neural networks (CNNs) have achieved significant success in image and object classification~\cite{simonyan2014very,he2016deep,szegedy2017inception,cen2019boosting,cen2019dictionary,ma2020mdfn},
the long-standing problem of partial occlusion remains a big challenge for the CNNs.
As a data-driven technique, the CNN-based classification usually requires massive labeled training data to model the diversity and alleviate overfitting in network training.
To train an occlusion-robust representation for the occluded images, \ie, to handle the occlusion from the occludee side, the training dataset has to be enlarged multiple times to cover the variations caused by the occlusion.
The collection and annotation of real data are, however, time-consuming and cost-expensive, especially for real occluded images.

In most practical applications, the training dataset usually contains much fewer occluded images (images are corrupted by the occluders) than clean images (images are not corrupted by the occluders), and sometimes, the available occluded images are irrelevant to the task-specific image classes.
In this paper, the occluders are referred to the image patches or objects that are irrelevant to the task, since handling the images with the objects occluded by the task-specific objects is usually regarded as a research topic of multi-label classification.
We consider the classification of generic occluded images under this application scenario with a further assumption that a small set of clean and occluded image pairs that are associated with the occluders (or similar occluders) in the occluded test images is available.
The images in the small set can either be included in or come from outside the original training set. 
Each of the clean and occluded image pairs is composed of a clean image and an occluded image, where the difference between the occluded image and the clean image is just the occluder\footnote{In practice, the non-occluded portion of the occluded image does not need to be identical to that of the clean image. The existence of small trivial difference in that portion can lead to almost the same result as the identical case, since the small trivial difference does not induce a large deviation in the deep feature space.}. 
This assumption can be easily satisfied in practice since only a small set is required.

Due to the limited number of the available occluded images, dealing with the occlusion from the occludee side in the above-mentioned application scenario is difficult. 
On the contrary, modeling an occluder usually requires much less training samples than modeling the variations of the occludees that are caused by the occluder.
Therefore, for the above-mentioned application scenario, we present a novel deep feature vector (DFV) augmentation approach in this paper to handle the occlusion in the deep feature space from the occluder side.

The DFV extracted by the CNNs is usually a holistic representation of the image, where the occlusion-related and the occlusion-unrelated elements are indistinguishable.
However, we observe that the difference vector (DV) between the DFVs of the occluded image and its original clean image is highly related to the occlusion.
Based on this observation and the following analysis, we propose to augment the DFVs in the fine-tuning stage with pseudo-DFVs that are generated by randomly adding the DVs extracted from a small set of clean and occluded image pairs.

\begin{figure}[t]
	\centering
	\includegraphics[width=0.7\linewidth]{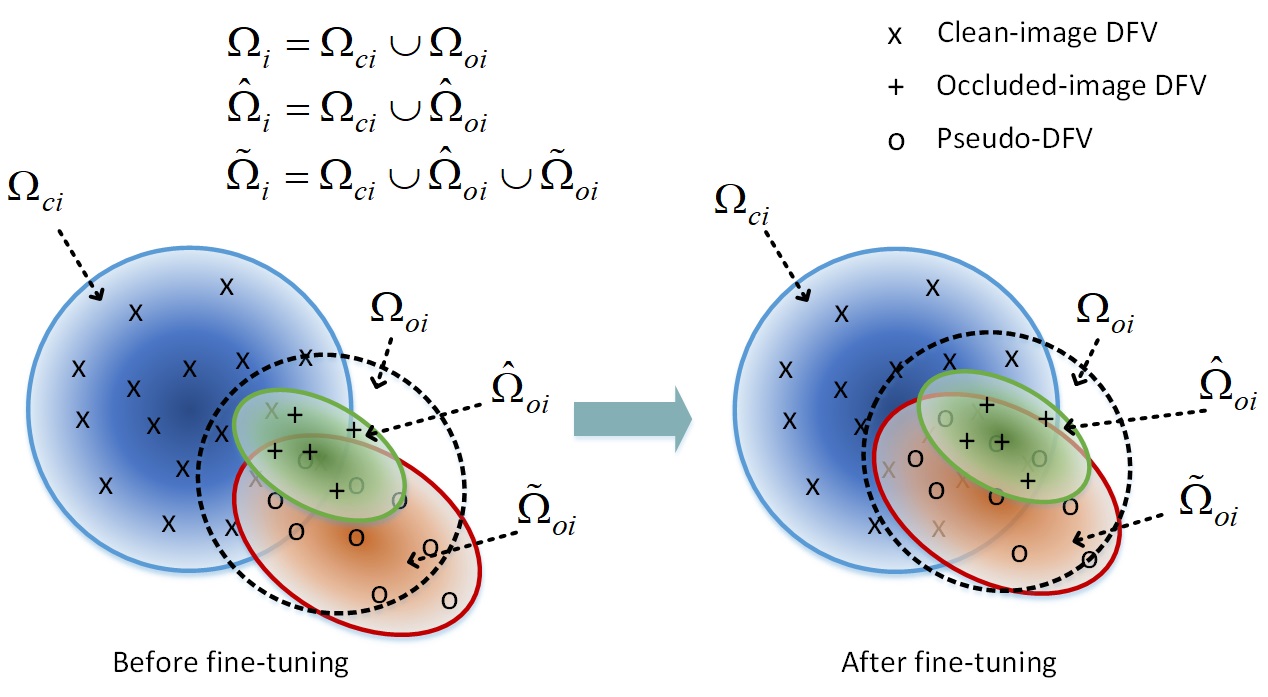}
	\vspace{-10pt}
	\caption{Illustration of the DFV augmentation concept. In the left diagram, without fine-tuning, the pseudo-DFVs can extend the classification boundary of the $i$th image class from the boundary of $\widehat{\Omega}_i$ to that of $\widetilde{\Omega}_i$. In the right diagram, by fine-tuning the CNN with the augmented DFVs, the area of $\Omega_{oi}$ outside $\widetilde{\Omega}_{i}$ (corresponding to false negative samples) and the area of $\widetilde{\Omega}_{oi}$ outside $\Omega_{oi}$ (corresponding to false positive samples) decrease. For simplicity, the shape variations of the subspaces are not shown. 
	\vspace{-10pt}
	}
	\label{fig:subspace_relation}
\end{figure}

Suppose we have a training set containing numerous clean images and a few occluded images.
A CNN well-trained on this training set performs a nonlinear mapping from the image space $\mathcal{X}$ to the deep feature space $\Omega$, $f:\mathcal{X}\longmapsto \Omega$.
The CNN maps the clean images and the occluded images of the $i$th image class to the deep feature subspaces: $\Omega_{ci}$ and $\Omega_{oi}$, respectively.
An ideal decision boundary to identify the images of the $i$th image class should coincide with the boundary of the subspace $\Omega_i=\Omega_{ci}\cup\Omega_{oi}$, as illustrated in the left diagram of Fig.~\ref{fig:subspace_relation}.
The classifier, however, can only learn the boundary of $\widehat{\Omega}_i=\Omega_{ci}\cup\widehat{\Omega}_{oi}$ from the training set\footnote{For simplicity, we suppose that the set of clean training images is sufficiently large such that the classifier can learn the boundary of $\Omega_{ci}$.}, where $\widehat{\Omega}_{oi}$ denotes the linear span of the DFVs of the occluded training images of the $i$th image class.
Apparently, $\widehat{\Omega}_{oi}$ is much smaller than $\Omega_{oi}$ since only a small subset of the occluded images is employed to train the networks.
As a result, the classification accuracy for the occluded images is very poor.

In our preliminary experiments (see Section~\ref{sec:correlation_dv} for details), we observed that the intra-pattern DVs associated with the similar occlusion patterns are close to each other on a low-dimensional manifold in the deep feature space.
Here, the occlusion pattern is defined as the occlusions with the same texture, shape, size, and location on the image.
This observation inspires us to generate a large set of pseudo-DFVs by randomly adding the DVs that are extracted from a small set of clean and occluded image pairs to the DFVs of the clean images.
The generated pseudo-DFVs have a high probability of falling into the subspace $\Omega_{oi}$ due to the closeness of the intra-pattern DVs to each other.
The linear span of the pseudo-DFVs $\widetilde{\Omega}_{oi}$ usually covers more space of $\Omega_{oi}$ than does $\widehat{\Omega}_{oi}$, as illustrated in the left diagram of Fig.~\ref{fig:subspace_relation}.

By augmenting the DFVs with the pseudo-DFVs in the training phase,
the learned decision boundary of the classifier, in theory, will be aligned with the boundary of $\widetilde{\Omega}_i=\Omega_{ci}\cup\widehat{\Omega}_{oi}\cup\widetilde{\Omega}_{oi}$.
If the back-propagation to the CNN is taken into account, the network parameters will be updated to represent the images in more compact subspaces, as illustrated in the right diagram of Fig.~\ref{fig:subspace_relation}, such that the classification accuracy for occluded images can be improved further.

The proposed approach is characterized by the following features.
\begin{enumerate}
	\vspace{-5pt}
	\setlength{\itemsep}{0pt}
	\item It is an end-to-end learning technique and can be directly applied to any pervasive CNNs without the need to modify the inference network structure.
	\item It requires only a small set of clean and occluded image pairs and thus is suited for practical applications.
	\item It is applicable to the classification for generic occluded images with the occluders of arbitrary shapes and textures.
	\item The model fine-tuned with the proposed approach does not require occlusion detection in inference and thus is universal for the classification of both the occluded images and the clean images. 
	\item It is validated on a large-scale general-purpose dataset with a large number of occluders.
\end{enumerate}
\vspace{-5pt}

The rest of this paper is organized as follows.
In section \ref{sec:related_works}, some related works are briefly reviewed.
Section \ref{sec:DF_augmentation} elaborates on the main observation on the relationship between the DVs and then presents the proposed deep feature augmentation approach.
Section \ref{sec:exp} presents the experimental results on both small- and large-scale datasets.
Finally, the paper is concluded in Section \ref{sec:conclusion}. The generated dataset and source code of the developed model will be available online.

\section{Related work} \label{sec:related_works}
\vspace{-5pt}
The related works on the classification of occluded images can be generally classified into two categories: with and without exploring the deep neural networks.

\vspace{-5pt}
\subsection{Without deep neural networks}
\vspace{-5pt}
\subsubsection{From occludee side}
In the early stage, only simple occluded objects on a clean background were investigated due to the limited computational power in that era.
The researchers focused on the occludee to develop manually designed occlusion-robust local features.
Polygonal approximations~\cite{liu1990partial}, circular arcs~\cite{ettinger1988large}, and boundary dominant points~\cite{tsang1994classification,zhang2003object} were employed to represent the occluded objects.
Later, to deal with the occlusion for the object with complex shape and texture, parts-based representation approaches were developed~\cite{huang1997object,ohba1997detectability}, which, in general, divide the object into different parts, and then encode the parts with robust local descriptors.
Unfortunately, these approaches only show effectiveness under well-controlled laboratory conditions with a clean background and viewed directly from the top of the objects, because how to manually design occlusion-robust features for general conditions has yet to be figured out.

Due to the difficulty in designing occlusion-robust features for generic images and the increasing interest in face recognition aroused by practical applications, researchers gradually moved their focus to occluded faces.
Following the route to tackle the occlusion from the occludee side and in the spirit of parts-based representation, many research attempts have been made.
Li \etal~\cite{li2001learning} proposed to learn spatially localized, parts-based subspace representation by imposing a localization constraint in the non-negative matrix factorization of face images.
Mart{\'\i}nez \cite{martinez2002recognizing} divided each face image into $6$ overlapped ellipse-shaped local regions which are analyzed separately, and each local part is modeled using a mixture of Gaussians.
Tan \etal~\cite{tan2005recognizing} extended the work in \cite{martinez2002recognizing} by using an equal size non-overlapping subblock partition and the self-organizing map (SOM) to learn the subspace that represented each individual. 
Kim \etal~\cite{kim2005effective} only employed locally salient information from important facial parts by imposing additional localization constraint in the process of computing ICA architecture I basis images.
\vspace{-7pt}
\subsubsection{From occluder side}
\vspace{-3pt}
A breakthrough in the field of occluded face classification was made by changing the focus from the occludee to the occluder.
In the seminal work, Wright \etal~\cite{wright2009robust} proposed a sparse representation-based classification (SRC) approach.
In SRC, the occluded face image was first coded as a sparse linear combination of the training samples and an occlusion dictionary. 
Then, the class is identified as that with the minimum reconstruction error.
Following Wright \etal's work, some researchers worked towards designing an advanced occlusion dictionary to model the contiguous occlusion.
Deng \etal~\cite{deng2012extended} employed the residual vectors between the clean samples and the occluded samples.
Yang \etal~\cite{yang2013gabor} introduced the Gabor feature into the SRC framework to compress the occlusion dictionary.
Ou \etal~\cite{ou2014robust} proposed to learn the occlusion dictionary from the data.
Yu \etal~\cite{yu2017discriminative} learned the occlusion auxiliary dictionary based on PCA.
Other researchers exploited various statistical properties and structural information of the occlusion errors.
Zhou \etal~\cite{zhou2009face} employed Markov Random Field to model the occlusion mask (error support) for contiguous occlusion.
Jia \etal~\cite{jia2012robust} modeled the occlusion error as a tree-structured sparse.
Li \etal~\cite{li2013structured} introduced a morphological graph model to describe the shape structure of the occlusion error and proposed a structured sparse error coding approach.
Yang \etal~\cite{yang2017nuclear}, Gao \etal~\cite{gao2017learning}, and Wu \etal~\cite{wu2018occluded} discovered the low-rank property of the occlusion error.

Although impressive improvements for occluded face recognition have been achieved, these works only show effectiveness on small-scale face datasets for aligned front-view faces with few occluders.
This weakness can be attributed to the sensitivity of the features exploited in these works, which are primarily the image pixels or a linear transformation of the image, to the cluttered background and many intra-class variations, such as pose changes and shape deformations.
Therefore, these works are difficult to be extended to the classification of the generic occluded images.

\subsection{With deep neural networks}
\subsubsection{Deep generative networks}
Deep network based approaches have attracted much attention in recent years.
Dealing with the occlusion in the deep feature space is, however, not a simple task because it is difficult to distinguish between the occlusion-related and the occlusion-unrelated deep features.
Consequently, inpainting with deep networks in the image space, which can be considered as a technique that addresses the occlusion problem from the occludee side, has attracted the interest of many researchers.

The deep generative models learned from the data are usually employed to estimate the original pixels of the occluded portions, \eg, the generative adversarial networks (GANs) were used in   \cite{pathak2016context,nguyen2017plug,yang2017high,yu2018generative}.
These approaches, however, primarily focus on pure inpainting, \ie, promoting the subject quality of the occluded images.
Pure inpainting without particular consideration on classification is not suited to the image classification task due to the inter-class information introduced by the filling patch, which is usually a combination (\eg~averaging) of multiple plausible candidates. 
For instance, the de-occluded mouth portion of an occluded face image, no matter how similar to a real mouth, involves the mouth information associated with multiple training individuals.
Moreover, these approaches need to know the shape and location of the occluder beforehand.

To avoid providing the prior knowledge about the structure of the occlusion, the technique of learning the occluder structure from data is developed under the generative model based framework.
Cheng \etal~\cite{cheng2015robust} employed an average relative difference between the activations of the occluded and the clean image as the criterion to discriminate between the corrupted and the non-corrupted elements of the DFVs for a stacked sparse denoising auto-encoder (SSDA) for face recognition.
Because of the holistic attribute of the DFVs, this approach only exhibits improvements for a few types of synthetic occlusions on a small-scale face dataset.

Zhao \etal~\cite{zhao2018robust} realized the deficiency in pure inpainting.
For the classification of occluded faces, they introduced an identity-based supervised CNN to provide extra guidance for the training of robust LSTM-autoencoders, which are employed to de-occlude face images.
However, numerous exactly matched face image pairs, each is composed of a clean face image and its corresponding occluded face image, are required to train the robust LSTM-autoencoders.
In practice, it is difficult to collect a large number of exactly matched face image pairs.

In inpainting, the deep generative models are usually sensitive to the type of the occludees, \ie, the model trained on one type of occludees (\eg~flowers) cannot be applied to other types of occludees (\eg~birds and animals).
The sensitivity to the type of the occludees hinders the generalization to other generic occluded images.

Recently, GAN-based data augmentation~\cite{DBLP:journals/corr/abs-1711-04340} has attracted great interest from many researchers. To the best of our knowledge, however, no report on successfully improving the classification or recognition of occluded images with the GAN-based data augmentation has been published.

\subsubsection{Other approaches}
Though difficult, handling the occlusion in the deep feature space has been investigated recently.
In~\cite{Song2019OcclusionRF}, a pairwise differential Siamese network (PDSN) was proposed to learn the correspondence between the occluded facial regions and the corrupted activations of the top convolution layer, and then, the occlusion-associated deep feature elements, which are indicated by a mask generated according to the learned correspondence, are discarded in classification.  
This approach, however, highly relies on well-aligned face images to build the correspondence and accurate detection of occlusion to generate the feature discarding mask.
This approach is not applicable to the classification of generic occluded images because object alignment and occlusion detection are still open challenges for the generic occluded images.

Attention-based methods have the potential to tackle the occlusion, but few of the researches focus on the classification of occluded images.
Xu \etal~\cite{juefei2016deepgender} introduced an attention-based method to progressively train the CNN focusing on particular regions of the face image, such as the regions around eyes, for the gender classification of occluded faces.
However, when the important portions of the image are occluded, this approach is unable to effectively discriminate the occluders from the image in those portions.
Therefore, the effectiveness is only shown in specific applications where the discriminative features lie on specific regions of the image.

According to the above analysis, we note that the impressive results have been achieved merely on occluded face datasets and a big challenge still remains on generic occluded images.
For the approaches applicable to large-scale occluded face datasets, either a huge number of exactly matched face image pairs or an accurate occlusion detector is required.
On the contrary, the proposed approach addresses the classification of occluded images on generic image datasets without a need for occlusion detector and by employing just a small set of clean and occluded image pairs.
In addition, the proposed approach is compatible with the data augmentation in the image space, since it conducts augmentation in the deep feature space.

\vspace{-5pt}
\section{Proposed approach}\label{sec:DF_augmentation}
\vspace{-5pt}
In this section, we will first describe our main observation and then introduce the proposed DFV augmentation approach.
\vspace{-5pt}
\subsection{Relationship between DVs} \label{sec:correlation_dv}
Let $\mathbf{v}_j$ denotes the DFV of the clean image and $\mathbf{\widehat{v}}_{jk}$ the DFV of the occluded image for a clean and occluded image pair of the $i$th image class and associated with the $k$th occlusion pattern.
The DFV is the output of the base CNN (excluding the prob layer and the last fully connected linear layer of the original CNN). 
The DV between the pair of DFVs, $\mathbf{\widehat{v}}_{jk}$ and $\mathbf{v}_j$ is given by
\begin{equation}\label{eq:dv_def}
\mathbf{d}_k =  \mathbf{\widehat{v}}_{jk} - \mathbf{v}_j,
\end{equation}
where, for simplicity, only the subscript $k$ is kept to indicate that the DV is associated with the $k$th occlusion pattern.

\begin{figure}[h]
	\centering
	\includegraphics[width=0.9\textwidth]{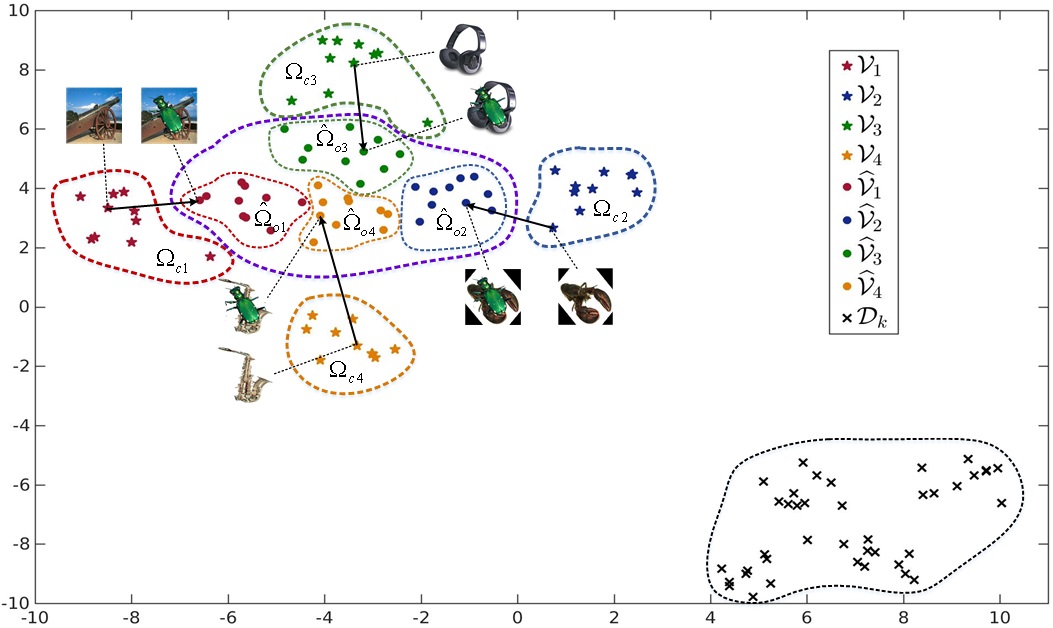}
	\caption{2-D illustration of DFVs and DVs for the synthetic dataset. For clarity, only one occlusion pattern, namely the $k$th occlusion pattern, is shown. The t-SNE algorithm~\cite{maaten2008visualizing}, which is an algorithm nonlinearly projecting the manifolds in the high-dimensional space to 2-D space for visualization, is employed to project the DFVs and DVs from $512$-D to $2$-D for visualization. Solid arrow lines are examples of the DVs in vector form. The corresponding images of the DFVs employed to generate the vector-form DVs are also shown. (better viewed in color) }
	\label{fig:dv_corr}
	\vspace{-10pt}
\end{figure}

In the following, a synthetic dataset is employed to illustrate the observations.
The synthetic dataset is composed of $4$ image classes each with $10$ clean images randomly drawn from the Caltech-101 dataset~\cite{fei2007learning} and $30$ occluded images synthesized with $20\%$ occlusion at the center.
Three occluders shown in Fig.~\ref{fig:caltech101_occlusion_example} are employed to generate the occluded images.
The examples of the clean images and occluded images are shown in Fig.~\ref{fig:dv_corr}. 
The $512$-dimensional DFVs are extracted with the ResNet18\_C model in Section~\ref{sec:caltech_occ_inc} (the ResNet18 network~\cite{he2016deep} fine-tuned on the Caltech-101 dataset with the classical training approach.) and the DVs are generated according to equation~(\ref{eq:dv_def}). 

Let $\mathcal{V}_i$ and $\widehat{\mathcal{V}}_i$ denote the set of the DFVs of the clean and the occluded images of the $i$th image class, respectively, and $\mathcal{D}_k$ the set of DVs associated with the $k$th occlusion pattern.
From Fig.~\ref{fig:dv_corr}, we can observe that the DFVs of the occluded images locate in a region enclosed by a purple dash curve and in that region the DFVs of each image class cluster together.
Such kind of distribution can be explained by the fact that the DFVs of the occluded images reflect the features of both the occludee and the occluder.  
The purple-curve enclosed region is actually a region on a low-dimensional manifold that is associated with the $k$th occlusion pattern.
From Fig.~\ref{fig:dv_corr}, we know that the occlusion causes the DFVs in $\Omega_{oi}$ moving into $\widehat{\Omega}_{oi}$.
The movements (as indicated by the DVs), though different for each individual DFV, have a certain kind of similarity because they are commonly associated with the same occlusion pattern and point to the positions close to each other on the low-dimensional manifold (the purple-curve enclosed region).
From Fig.~\ref{fig:dv_corr}, we can observe that the DVs (black 'x' marks) locate in a small area enclosed by a black dash curve.
This indicates that \textbf{the intra-pattern DVs (associated with the same occlusion pattern) are close to each other on a low-dimensional manifold in the deep feature space}\footnote{Strictly speaking, in practice, for each occlusion-pattern, this observation may not be met by all of the intra-pattern DVs but can be satisfied by the majority of them for a well-trained CNN. See Section S.I of the supplementary document for the evidence on the whole Caltech-101 dataset.}.
Note that the intra-pattern DVs can be generated with the DFVs acquired from any image classes.

The above observation implies that if an intra-pattern DV is added to the DFV of a clean image, it has a high probability that the generated pseudo-DFV locates close to the real DFVs of the occluded images associated with the given occlusion pattern.
Therefore, we can extend the span of the occluded images in the deep feature space with the pseudo-DFVs, \ie~enlarge $\widehat{\Omega}_{oi}$.

The generic model to generate the pseudo-DFV $\widetilde{\mathbf{v}}_{ik}$, which belongs to the $i$th image class and is associated with the $k$th occlusion pattern, can be written as
\begin{equation}\label{eq:psuedo_dfv}
\widetilde{\mathbf{v}}_{ik} = \mathbf{v}_i + \beta\mathbf{d}_k,
\end{equation}
where $\beta$ is the addition weight.
It is worth noting that in equation (\ref{eq:psuedo_dfv}), $\mathbf{d}_k$ can be generated from the pair of DFVs associated with any image class either the same as or distinct from the $i$th image class.

\begin{figure*}
	\centering
    \includegraphics[width=0.9\linewidth]{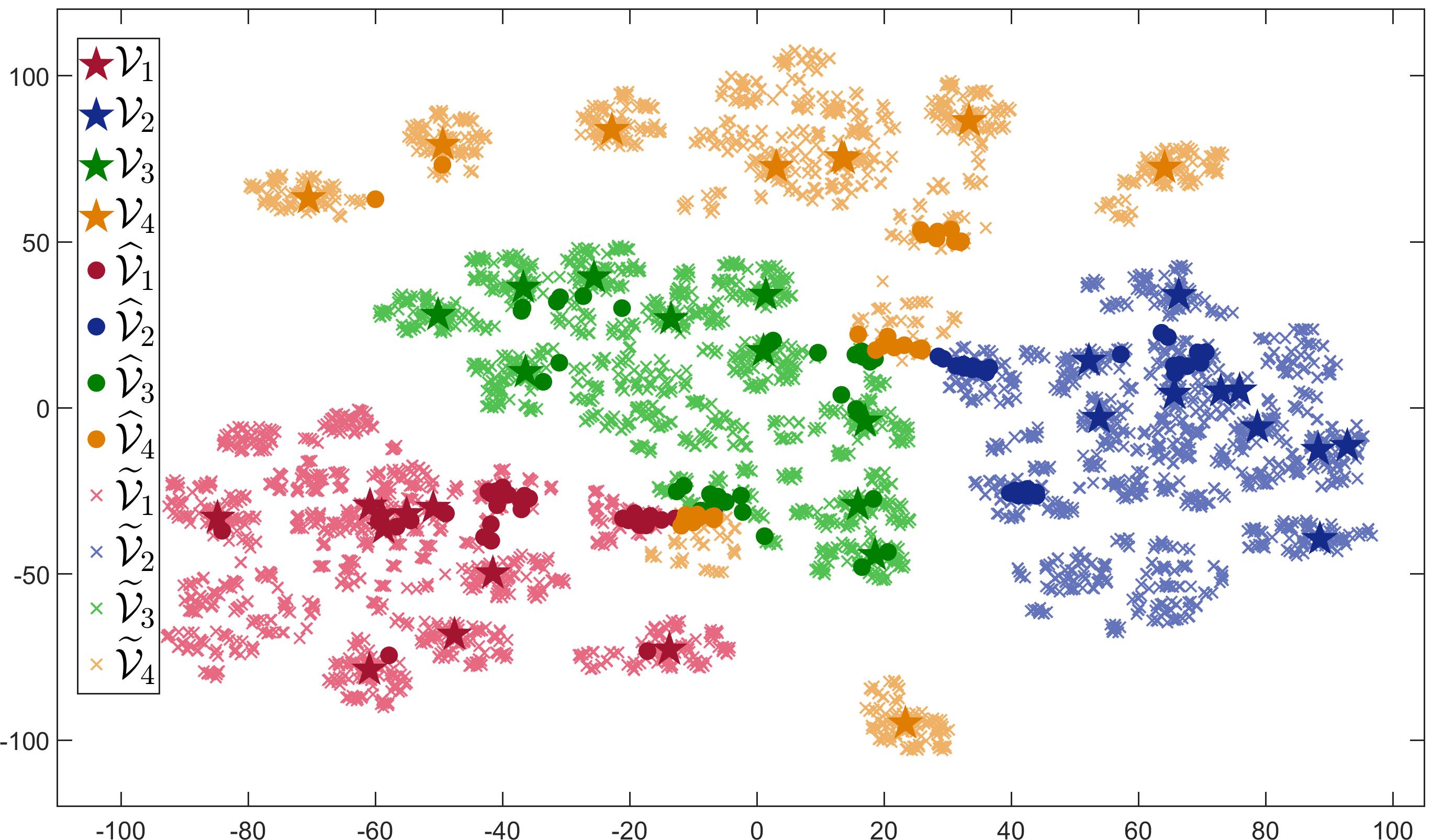}
	\caption{The 2-D illustration of DFVs and pseudo-DFVs for the synthetic example. $\mathcal{V}_i$: the set of the clean-image DFVs of the $i$th image class; $\widehat{\mathcal{V}}_i$: the set of the occluded-image DFVs of the $i$th image class; $\widetilde{\mathcal{V}}_i$: the set of the pseudo-DFVs of the $i$th image class.
		The t-SNE~\cite{maaten2008visualizing}	is employed to project the DFVs and DVs from $512$-D to $2$-D for visualization (better viewed in color).
	}
	\label{fig:pseudoDFVs}
	\vspace{-10pt}
\end{figure*}

To verify the role of the pseudo-DFVs, the real DFVs and the pseudo-DFVs generated with $\beta=1$ are plotted in Fig.~\ref{fig:pseudoDFVs} for the synthetic dataset.
For each real DFV, the pseudo-DFVs are yielded by using $117$ DVs (all of the DVs other than three associated with the real DFV).
In Fig.~\ref{fig:pseudoDFVs}, $\widetilde{\mathcal{V}}_i$ denotes the pseudo-DFV set of the $i$th image class.
From Fig.~\ref{fig:pseudoDFVs}, we have the following observations.
\begin{enumerate}
	\setlength{\itemsep}{0pt}
	\item The pseudo-DFVs greatly enlarge the number and the occupied space of the DFVs of the occluded images. In Fig.~\ref{fig:pseudoDFVs}, the number of the DFVs of the occluded images for each image class is $30$, while the number of pseudo-DFVs for each image class is $1170$. It is evident that the pseudo-DFVs occupy much larger space than the DFVs of the occluded images.
	\item The pseudo-DFVs that are generated from the DFVs of the $i$th image class fall into $\Omega_i$, the subspace of the $i$th image class.
	In Fig.~\ref{fig:pseudoDFVs}, we can see that for each image class, the vectors in $\mathcal{V}_i$ and $\widehat{\mathcal{V}}_i$ are surrounded by the vectors in $\widetilde{\mathcal{V}}_i$ and the vectors in $\widetilde{\mathcal{V}}_i$ can be easily separated from the DFVs and pseudo-DFVs of other three image classes.			
	\item The space occupied by the pseudo-DFVs is highly overlapped with the space covered by the DFVs of the occluded images of the same image class.	
	For instance, in Fig.~\ref{fig:pseudoDFVs}, the area occupied by $\widetilde{\mathcal{V}}_1$ covers almost the whole area occupied by $\widehat{\mathcal{V}}_1$. 
\end{enumerate}

The above observations manifest the analysis for the role of the augmented DFVs in Section \ref{sec:intro}.

\vspace{-5pt}
\subsection{Deep feature augmentation}
\begin{figure*}
	\centering
	\includegraphics[width=0.9\linewidth]{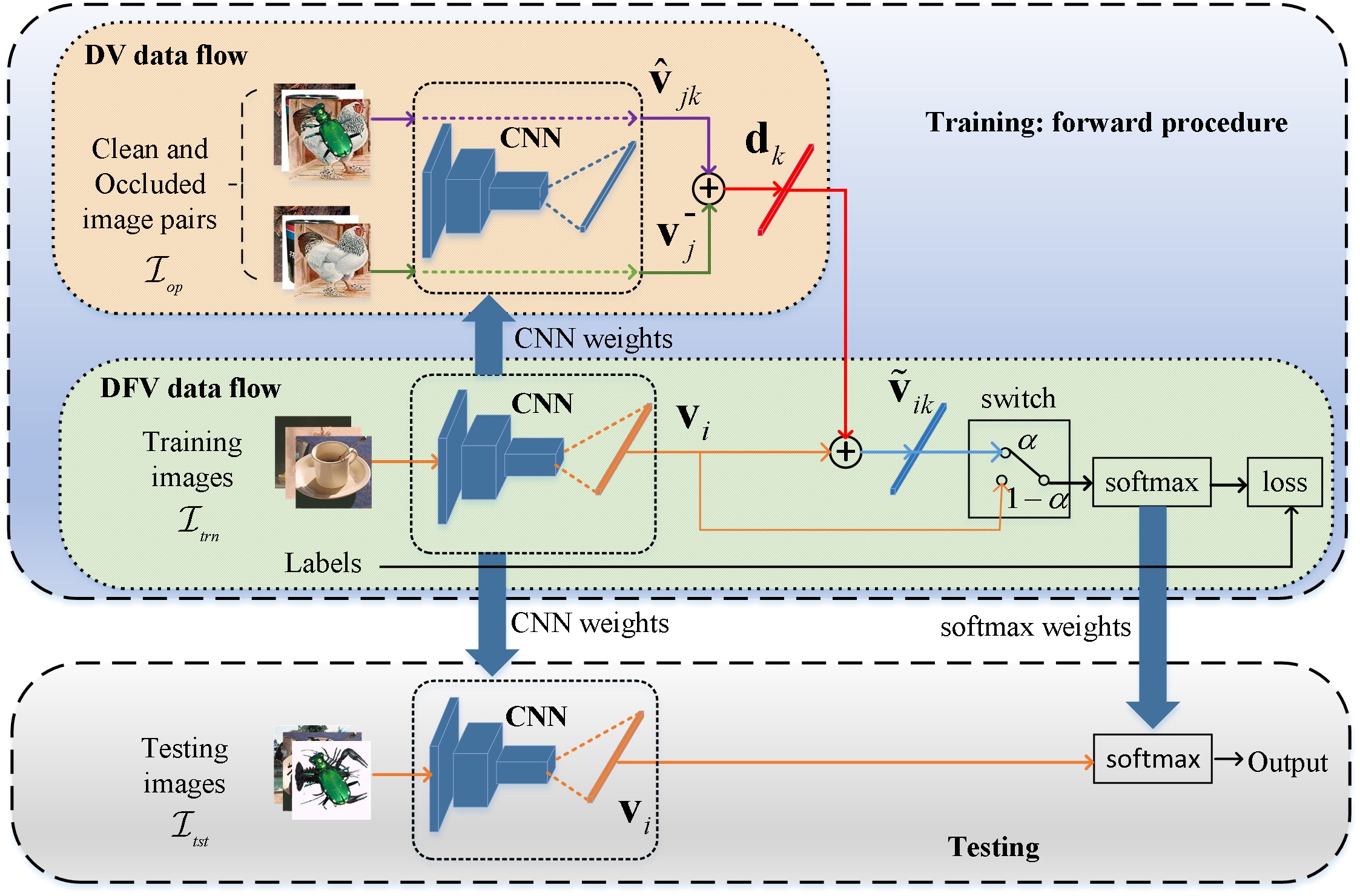}
	\caption{Illustration of the forward training procedure for the proposed deep feature augmentation approach and the testing procedure.
	}
	\label{fig:sys_diagram}
	\vspace{-10pt}
\end{figure*}
Before elaborating on the proposed approach, we first define some notations on the image sets used in the remaining part of this paper.
\begin{itemize}
	\setlength{\itemsep}{0pt}
	\item $\mathcal{I}_{trn}$: the set of training images fed into the DFV data flow in Fig.~\ref{fig:sys_diagram}. The subsets of clean images and occluded images are denoted by $\mathcal{I}_{trn\_c}$ and $\mathcal{I}_{trn\_o}$, respectively.
	\item $\mathcal{I}_{tst}$: the set of test images used for evaluation. The subsets of clean images and occluded images are denoted by $\mathcal{I}_{tst\_c}$ and $\mathcal{I}_{tst\_o}$, respectively.
	\item $\mathcal{I}_{op}$: the set of clean and occluded image pairs fed into the DV data flow in Fig.~\ref{fig:sys_diagram}. The sets of clean images and occluded images in $\mathcal{I}_{op}$ are denoted by $\mathcal{I}_{op\_c}$ and $\mathcal{I}_{op\_o}$, respectively.
	\vspace{-5pt}
\end{itemize}

In the following experiments on the synthetic occluded images, the clean images are referred to as the original images of the dataset and the occluded images as the synthetic occluded images.

The proposed deep feature augmentation approach is illustrated in Fig.~\ref{fig:sys_diagram}.
The deep feature augmentation is only conducted in the training stage.
There are two data flows, DV data flow and DFV data flow, in the forward procedure.
The base CNNs in the two data flows have the same structure.
The DV data flow is used to extract the DVs from $\mathcal{I}_{op}$.
In the DFV data flow, the training images are first fed into the base CNN to extract the DFVs, and then, the DVs from the DV data flow are randomly added to the DFVs according to equation (\ref{eq:psuedo_dfv}) to yield the pseudo-DFVs.
The pseudo-DFVs and original DFVs are sent to the softmax layer with the passing-through probabilities of $\alpha$ and $1-\alpha$, respectively.
Here, $\alpha$ is the control parameter of the switch in Fig.~\ref{fig:sys_diagram}.
The switch has the probability of $\alpha$ to connect to the pseudo-DFV data path (the upper path).
The function of the switch is to ensure the CNN can be trained for the classification of both clean images and occluded images.
The backpropagation is conducted merely on the DFV data flow. 

After training with the proposed approach, the trained CNN model is treated as a normal CNN model in the testing stage, as illustrated in Fig.~\ref{fig:sys_diagram}. 

\vspace{-5pt}
\subsection{Implementation consideration}
\vspace{-4pt}
In our implementation, the weights of the base CNN in the DV data flow is only updated at the start of each training epoch with the weights of the base CNN in the DFV data flow.

In Fig.~\ref{fig:sys_diagram}, the DFVs can be thought of as a special type of pseudo-DFVs generated by adding a zero DV to the DFVs.
Therefore, in our implementation, instead of employing the switch block in Fig.~\ref{fig:sys_diagram}, we randomly insert the zero vectors into the sequence of DVs with a probability of $1-\alpha$ and send the pseudo-DFVs (including the DFVs as the special pseudo-DFVs) directly to the softmax block. 
This implementation facilitates the backpropagation.

Since the observations mentioned in Section \ref{sec:correlation_dv} about the DVs are only shown by pre-trained CNN models, we simply apply the proposed deep feature augmentation approach to fine-tune the pre-trained CNNs. 

In deep network training, lots of training tricks can be applied to improve performance. Some popular tricks include the size
of minibatch, data augmentation, the number of training epochs,
gradient descending algorithm, the number of fine-tuning layers,
and the learning rate updating schedule. 
For a fair comparison, in the following experiments, the same training tricks are adopted for both the training with the classical approach and the fine-tuning with the proposed approach.

\vspace{-10pt}
\section{Experimental results}\label{sec:exp} 
\vspace{-5pt}
In this section, we present extensive experiments on a small-scale dataset (Caltech-101 \cite{fei2007learning}), a real occluded face dataset (AR \cite{martinez1998ar}), and a large-scale dataset (ImageNet).
Two types of networks are evaluated in the experiments,  ResNet \cite{he2016deep} and VGG \cite{simonyan2014very}.
The ResNet networks are evaluated on all the datasets to validate the proposed approach;
the VGG networks are only evaluated on the ImageNet dataset to manifest the generalization of the proposed approach to different network structures. 

The Caltech-101 dataset contains images of objects grouped into 101 image classes.
The Caltech-101 dataset excluding the "background" class is adopted to evaluate the basic properties of the proposed approach.
The AR dataset contains face images with different facial expressions, illumination conditions, and occlusions.
As in \cite{wright2009robust}, a subset face crops of the AR dataset that contains 50 male and 50 female subjects are adopted for evaluation. 
The ImageNet is a comprehensive large-scale database.
A subset of ImageNet database, the ImageNet Large-Scale Visual Recognition Challenge 2012 (ILSVRC2012) \cite{ILSVRC15} classification dataset consisting of 1000 image classes, is adopted for evaluation.
For simplicity, in the following experiments, for the evaluation on the synthetic occluded datasets, the images in the original dataset are treated as the clean images.

Two types of training sets, occlusion-exclusive training set, \ie, $\mathcal{I}_{op}\cap \mathcal{I}_{trn} = \emptyset$, and occlusion-inclusive training set, \ie, $\mathcal{I}_{op}\subset \mathcal{I}_{trn}$, are set up for the following experiments. 
The occlusion-exclusive training set corresponds to the application situation that only a set of clean and occluded image pairs unrelated to any task-specific image classes can be acquired. 
For instance, in the application of recognizing animals, one may only have some clean and occluded flower image pairs in hand.
The occlusion-inclusive training set corresponds to the scenario that a set of clean and occluded image pairs related to the task-specific image classes are available for network training. 

Due to the lack of publicly available datasets with a large number of real occluded images, to simplify the evaluation, without loss of generality, except for the experiment on the AR dataset, the synthetic occluded images are employed for evaluation.
The synthetic occluded images are generated with two steps.
First, the clean images are resized to $224\times 224$, which is the input size of the base CNNs used in the experiments.
Specifically, for the Caltech-101 dataset, the image is directly resized to $224\times 224$, while, for the ILSVRC2012 dataset, the clean image is a center $224\times 224$ crop from the resized image with shorter side equal to 256.
Then, the occlusion patches are scaled according to the occlusion ratio and superimposed on the resized clean images.
In addition, in all the following experiments, the data augmentation is only applied to the DFV data flow.

\subsection{Caltech-101}\label{sec:caltech101}
\begin{figure*}
	\centering
	\includegraphics[width=0.9\linewidth]{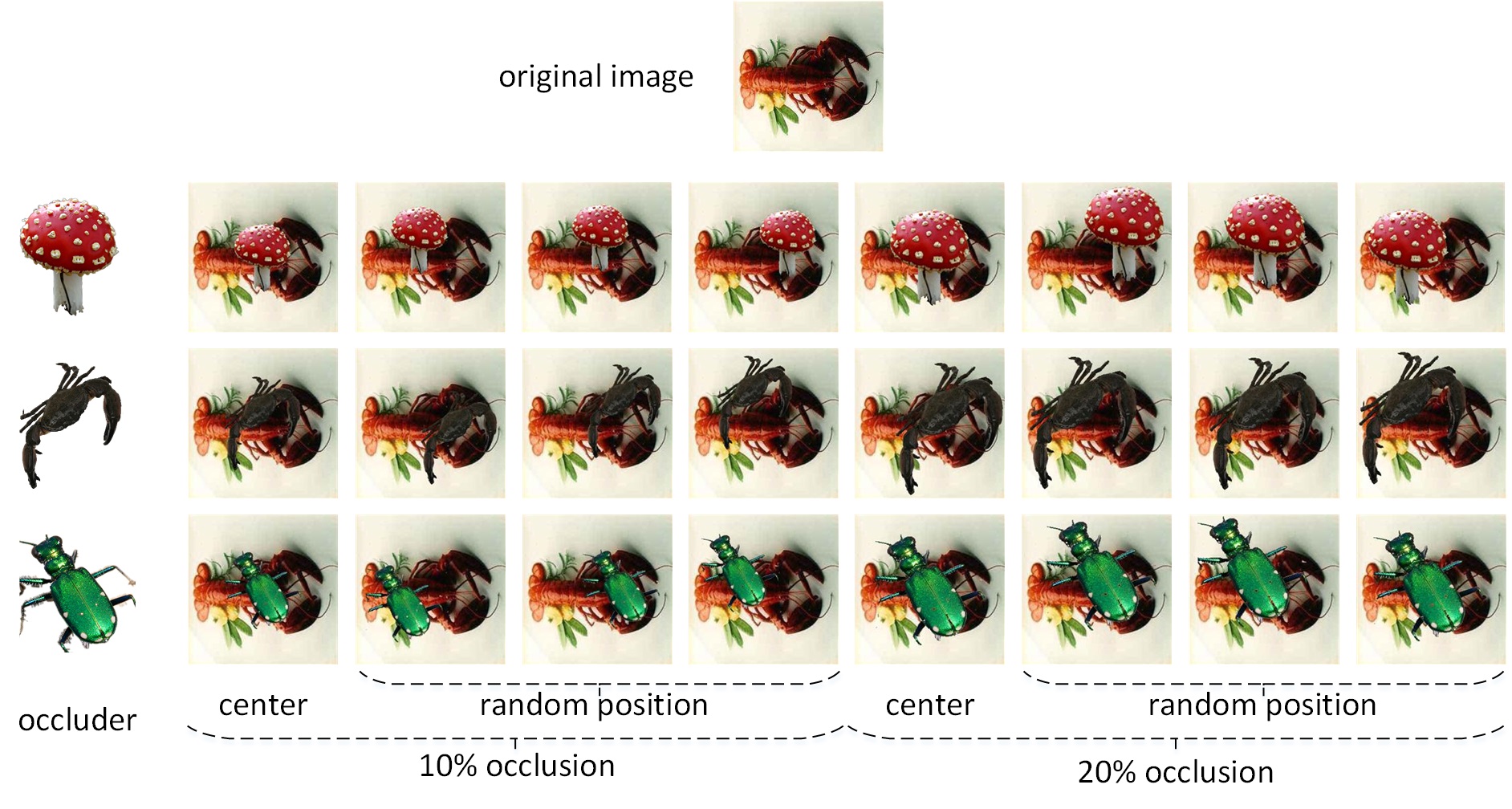}
	\vspace{-10pt}
	\caption{Three Occluders and example images for each occlusion pattern used for the evaluations on the Caltech-101 dataset.
	}
	\label{fig:caltech101_occlusion_example}
	\vspace{-10pt}
\end{figure*}

The experiments in this section are designed to evaluate the basic properties of the proposed approach on the Caltech-101 dataset.
In the experiments, all of the training and test images are resized directly to $224\times 224$ and the ResNet18 network~\cite{he2016deep} pre-trained on the ILSVRC 2012 dataset is adopted as the base CNN for fine-tuning.
$24$ occlusion patterns in total, three occluders each with two occlusion ratios ($10\%$ and $20\%$) and four occlusion positions (the center and three random positions),  are adopted to synthesize the occluded images.
The occluders and example images for each occlusion pattern are shown in Fig.~\ref{fig:caltech101_occlusion_example}.
The data augmentations of horizontal flip and random affine transformation are applied to train all of the models. 

\subsubsection{Occlusion-exclusive training set} \label{sec:caltech_occ_exc}
In this experiment, we evaluate the proposed approach on the occlusion-exclusive training set.
Three $\mathcal{I}_{op}$ settings,  $\mathcal{I}_{op}^{\text{Full}}$, $\mathcal{I}_{op}^{\text{10C}}$, and $\mathcal{I}_{op}^{\text{20C}}$, are assessed to show the properties of the proposed approach.
$\mathcal{I}_{op}^{\text{Full}}$ contains the same occlusion patterns as $\mathcal{I}_{tst}$; $\mathcal{I}_{op}^{\text{10C}}$ contains the occlusion patterns only with $10\%$ occlusion at the center of the image; $\mathcal{I}_{op}^{\text{20C}}$ contains the occlusion patterns only with $20\%$ occlusion at the center of the image.

To construct the occlusion-exclusive training set, we split the dataset as follows.
$91$ image classes with the names from "accordion" to "sunflower" in alphabet order are employed for $\mathcal{I}_{trn}$ and $\mathcal{I}_{tst}$ and the remaining $10$ image classes for $\mathcal{I}_{op}$.
The images of $\mathcal{I}_{trn}$ are randomly drawn from each image class with $30$ images per image class and the clean images of $\mathcal{I}_{tst}$ from the remaining images of each image class with a maximum of $50$ images per image class.
For each image class in $\mathcal{I}_{op}$, we randomly drawn $5$ images as the clean images and then corrupt them with the occlusion patterns to generate the occluded images. 
Eventually, we have $50$ clean and occluded image pairs for each occlusion pattern. 
The settings are summarized in Table \ref{tab:caltech101_excl_settings}.
\begin{table}
	\renewcommand{\arraystretch}{0.9}
	\centering
	\fontsize{8}{8}\selectfont
	\caption{The settings for the image sets used in the experiment on the Caltech-101 dataset for the occlusion-exclusive training set.}
	\begin{tabular}{c|c|c|c}
		\hline
		\multirow{2}{*}{Image set} & \multirow{2}{*}{Image classes} & \multicolumn{2}{c}{Images per class} \\ \cline{3-4} 
		&  & Clean & Occluded \\ \hline
		$\mathcal{I}_{trn}$ & 91 & 30 & - \\ \hline
		$\mathcal{I}_{op}^{\text{Full}}$ & 10 & 5 & 120 \\ \hline
		$\mathcal{I}_{op}^{\text{10C}}$ & 10 & 5 & 15 \\ \hline
		$\mathcal{I}_{op}^{\text{20C}}$ & 10 & 5 & 15 \\ \hline
		$\mathcal{I}_{tst}$ & 91 & $\leq $50 & $\leq $1200 \\ \hline
	\end{tabular}	
	\label{tab:caltech101_excl_settings}
	\vspace{-10pt}
\end{table}

The following four models, fine-tuned on $\mathcal{I}_{trn}$, are evaluated on $\mathcal{I}_{tst}$.
\begin{itemize}
	\setlength{\itemsep}{0pt}
	\item ResNet18: ResNet18 network is fine-tuned with classical training approach.
	\item ResNet18-Full: ResNet18 network is fine-tuned with the proposed approach and $\mathcal{I}_{op}^{\text{Full}}$ is adopted in training.
	\item ResNet18-10C: ResNet18 network is fine-tuned with the proposed approach and $\mathcal{I}_{op}^{\text{10C}}$ is adopted in training.
	\item ResNet18-20C: ResNet18 network is fine-tuned with the proposed approach and $\mathcal{I}_{op}^{\text{20C}}$ is adopted in training.	
\end{itemize}

Here, the classical training approach refers to the broadly used training approach for general image classification without exploiting the pseudo-DFVs. 

First, we employ ResNet18-Full as an example to show the variation of the classification accuracy \wrt~the hyperparameters $\alpha$ and $\beta$.
\begin{figure}
	\centering
		  \subfloat[]{
		\begin{minipage}[c][]{
				0.49\textwidth}
			\centering
			\includegraphics[width=1\textwidth]{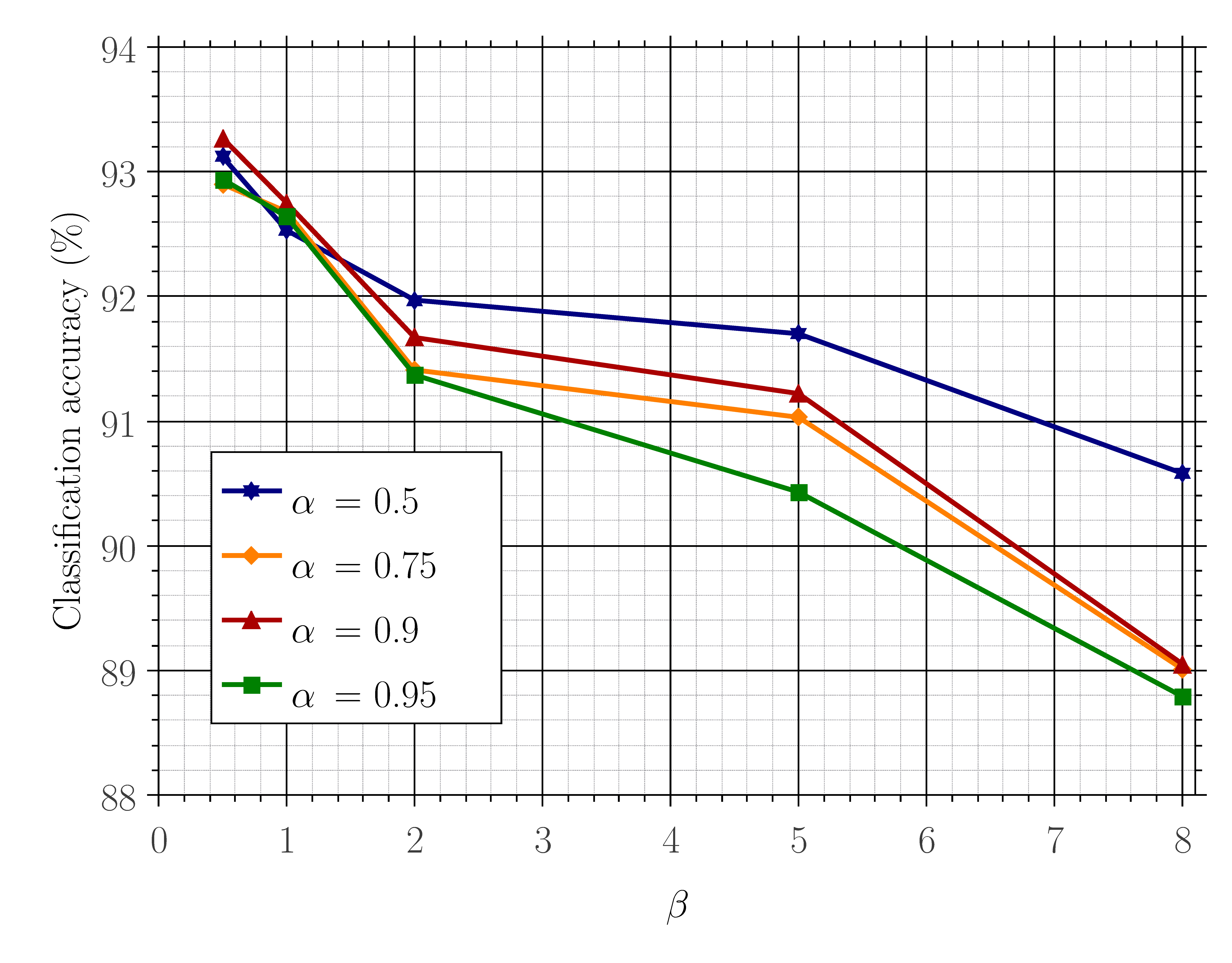}\vspace{-10pt}
	\end{minipage}}
	\hfill
	\subfloat[]{
		\begin{minipage}[c][]{
				0.49\textwidth}
			\centering
			\includegraphics[width=1\textwidth]{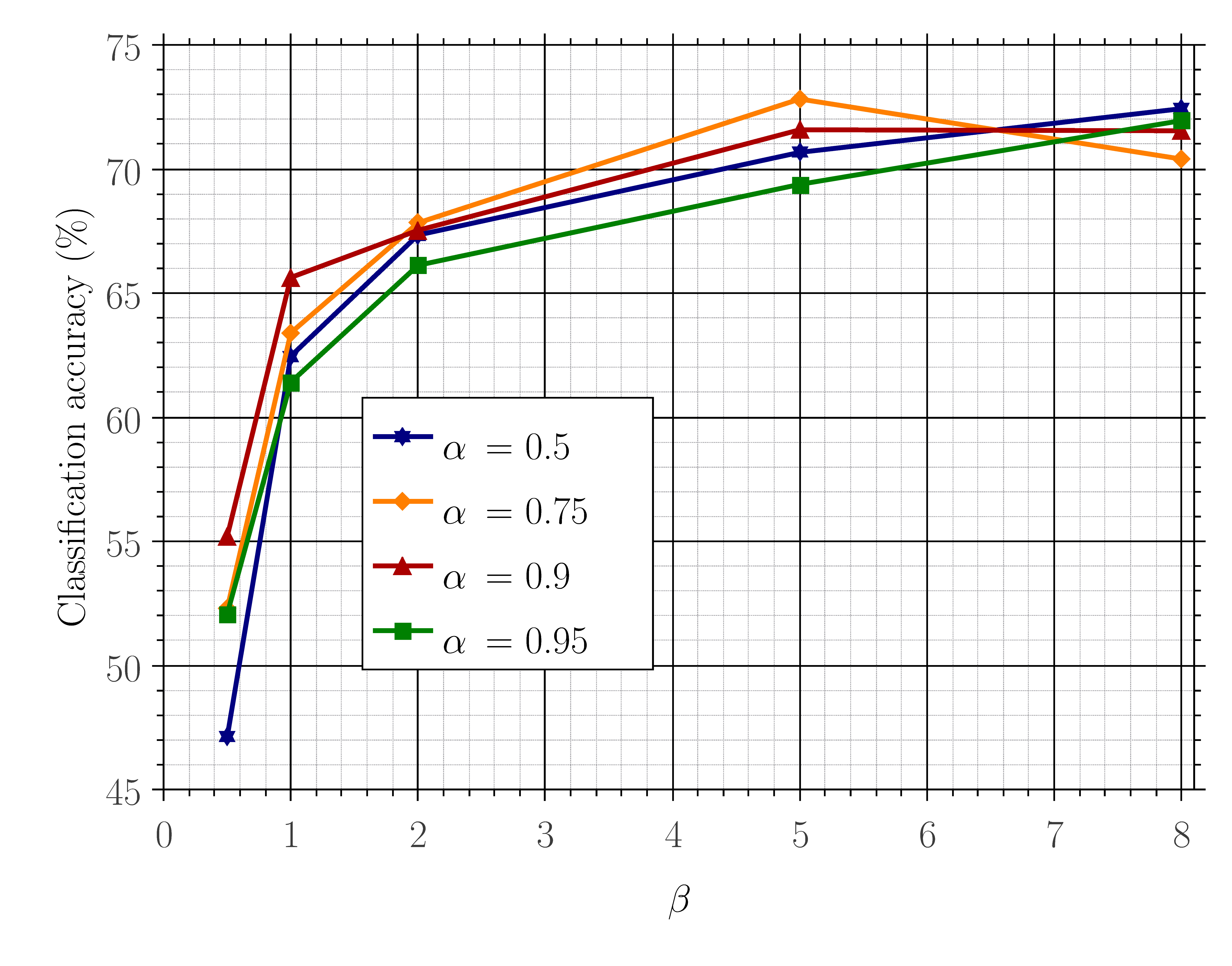}\vspace{-10pt}
	\end{minipage}}	\vspace{-10pt}
	\caption{The classification accuracy vs. $\beta$ for (a) the set of clean test images, $\mathcal{I}_{tst\_c}$, and (b) the set of occluded test images, $\mathcal{I}_{tst\_o}$, for the ResNet18-Full model. (better viewed in color) 
	}
	\label{fig:alpha_accuracy}
	\vspace{-10pt}
\end{figure}

When $\beta \gg 1$, $\beta \mathbf{d}_k$ has a large deviation from $\mathbf{d}_k$, and it will cause some of the pseudo-DFVs falling into the clean-image subspaces of other image classes, and thus, the performance for the clean images is undermined.
We can see that in Fig.~\ref{fig:alpha_accuracy}~(a), the classification accuracy for $\mathcal{I}_{tst\_c}$ decreases significantly when $\beta \gg 1$.
When $\beta<1$, the pseudo-DFVs mainly stay inside the occluded-image subspaces, thus having a small impact on the classification of clean images but a large influence on the classification of occluded images.
As shown in Fig.~\ref{fig:alpha_accuracy}~(b), for $\beta<1$, the classification accuracy for $\mathcal{I}_{tst\_o}$ rises rapidly with $\beta$ but that for $\mathcal{I}_{tst\_c}$ changes pretty small (less than $0.6\%$).

The hyperparameter $\alpha$ indicates the percentage of the pseudo-DFVs involved in each minibatch.
To optimize a pre-trained model for an old and a new task simultaneously, much more new training samples are usually required than the old training samples to drift the parameters to new optimal values.
For a model pre-trained on the clean images, 
the classification of the occluded images is apparently a new task. 
For a small $\beta$, the classification of the clean images can be viewed as the old task because the pseudo-DFVs have little influence on the clean-image subspaces \footnote{Though ResNet18 is pre-trained on the ILSVRC2012 dataset, the DFVs extracted from the Caltech-101 dataset approximately cluster together for each image class. Hence, the classification of the clean images can be regarded as the old task.}. 
Therefore, for a small $\beta$, with the increase of $\alpha$, significant improvements are shown in Fig.~\ref{fig:alpha_accuracy}~(b) and small variations in Fig.~\ref{fig:alpha_accuracy}~(a).
While, for a large $\beta$, the classification of the clean images should also be considered as a new task because some of the pseudo-DFVs fall into the clean-image subspaces of other image classes.
In such a situation, the variations of the performances \wrt~$\alpha$ in Fig.~\ref{fig:alpha_accuracy}~(a) and (b) are inverse, since the classifications of the clean and the occluded images both are new tasks. 
However, in Fig.~\ref{fig:alpha_accuracy}, we can also observe that the performance for $\alpha=0.95$ is worse than that for $\alpha=0.9$ for most of $\beta$'s.
The reason is that if $\alpha$ is close to $1$, \ie~the percentage of real DFVs in
each minibatch becomes very small, the model will be unable to be optimized for the clean images
and thus, the classification accuracy declines. 

In practice, the setting of $\alpha$ and $\beta$ can be determined accordingly depending on the application requirement, such as whether the application focus on the classification of the occluded images or not.
In most applications, the classification of the clean images is the major target.
From Fig.~\ref{fig:alpha_accuracy}, we can know that at $\beta=1$, the loss in the performance for the clean image is very small and a turning point of the performance increment for the occluded images is shown. 
Furthermore, the non-one $\beta$ will introduce unrealistic DVs into the pseudo-DFVs, which may lead to unexpected results in some applications.
For $\beta=1$, in Fig.~\ref{fig:alpha_accuracy}, $\alpha=0.9$ shows the best performance for both the occluded and the clean images. 
Therefore, $\alpha=0.9$ and $\beta=1$ are adopted in the following experiment.

Then, we list the classification results of all four models for the clean images and the occluded images in Table \ref{tab:cltch101_exclusive}.
\begin{table}[h]
	\renewcommand{\arraystretch}{0.9}	
	\caption{Comparison of classification accuracy (\%) for the occlusion-exclusive training set on the Caltech-101 dataset.}
	\centering
	\fontsize{8}{8}\selectfont
	\begin{tabular}{c|c|c|c|c|c|c}
		\hline
		Model&\multirow{2}{*}{\begin{tabular}[c]{@{}c@{}}Clean\\images\end{tabular}}  &\multicolumn{2}{c|}{$20\%$ occlusion images} & \multicolumn{2}{c|}{$10\%$ occlusion images} & \multirow{2}{*}{\begin{tabular}[c]{@{}c@{}}Avg. over\\ occlusion\end{tabular}}\\
		\cline{3-6}
		&  & center & random & center &random & \\
		\hline
		ResNet18&93.05 & 22.66 & 30.59 &47.91 & 61.74& 43.44 \\	
		\hline
		ResNet18-Full&92.75 & 46.87& 57.3 &68.76 & 79.18& 65.63 \\		
		\hline
		ResNet18-10C&91.92 & 39.14 & 49.91 &65.7 & 75.72& 60.22 \\	
		\hline
		ResNet18-20C&92.37 & 47.42 & 57.99 &65.63 & 76.28& 64.48 \\	
		\hline
	\end{tabular}
	\label{tab:cltch101_exclusive}
	\vspace{-10pt}
\end{table}

We can observe that the proposed approach significantly improves the performance for the occluded images (a minimum of $16.78\%$ increase in accuracy averaged on all of the occlusion patterns) with slight degradation for the clean images (a maximum of $1.13\%$ decrease in accuracy).
We also note that, though trained with subsets of the occlusion patterns in $\mathcal{I}_{tst}$ ($20\%$ and $10\%$ occlusion positioned at the center of the images for ResNet18-20C and ResNet18-10C, respectively), both of ResNet18-20C and ResNet18-10C achieve better classification accuracies for all of the occlusion patterns than ResNet18.
This indicates that the DFV can be augmented with similar occlusion patterns, such as the occlusion patterns with the same texture but distinct occlusion ratios and positions.

From Table~\ref{tab:cltch101_exclusive}, we also learn that the average performance of ResNet18-20C is close to that of ResNet18-Full and much higher than that of ResNet18-10C.
This can be understood as follows.
The DVs associated with the centered $20\%$ occlusion have large diversity, and thus, their linear span covers most portion of the linear span of the DVs associated with the remaining occlusion patterns.
Therefore, ResNet18-20C achieves the classification accuracy similar to that of ResNet18-Full.
The DVs associated with the centered $10\%$ occlusion have small diversity, and thus, their linear span only covers a small portion of the linear span of the DVs associated with the remaining occlusion patterns, which results in the lower performance for ResNet18-10C.

\subsubsection{Occlusion-inclusive training set}\label{sec:caltech_occ_inc}
In this experiment, the performance of the proposed approach on the occlusion-inclusive training set is evaluated.
To assess the influence of the occluded images in the training set, we adopted two types of training sets, $\mathcal{I}_{trn}$ and $\mathcal{I}_{trn\_c}$.
$\mathcal{I}_{trn\_c}$ is the clean image subset of $\mathcal{I}_{trn}$.
The images of $\mathcal{I}_{trn\_c}$ are randomly drawn from $101$ image classes each with $30$ images and the clean images of $\mathcal{I}_{tst}$ are from the remaining images of each image class with a maximum of $50$ images per image class.
$\mathcal{I}_{op}$ is composed of 10 image classes randomly drawn from 101 image classes and each with 5 clean images randomly drawn from $\mathcal{I}_{trn\_c}$.
So, $\mathcal{I}_{op\_c}\subset \mathcal{I}_{trn\_c}$.
The occluded images of $\mathcal{I}_{tst}$ and $\mathcal{I}_{op}$ are synthesized with all 24 occlusion patterns.
The training set $\mathcal{I}_{trn}$ is composed of $\mathcal{I}_{trn\_c}$ and all of the occluded images of $\mathcal{I}_{op}$, \ie,  $\mathcal{I}_{trn} = \mathcal{I}_{trn\_c}\cup \mathcal{I}_{op\_o}$.
The settings for the image sets are summarized in Table \ref{tab:caltech101_incl_settings}.
\begin{table}[h]
	\renewcommand{\arraystretch}{0.9}
	\centering
	\fontsize{8}{8}\selectfont
	\caption{The settings for the image sets used in the experiment on the Caltech-101 dataset for the occlusion-inclusive training set.}
	\begin{tabular}{c|c|c|c|c}
		\hline
		\multirow{2}{*}{image set} & \multicolumn{2}{c|}{Clean images} & \multicolumn{2}{c}{Occluded images} \\ \cline{2-5} 
		& Classes & Images per class & Classes & Images per class \\ \hline
		$\mathcal{I}_{trn}$ & 101 & 30 & 10 & 240 \\ \hline
		$\mathcal{I}_{trn\_c}$ & 101 & 30 & - & - \\ \hline
		$\mathcal{I}_{op}$ & 10 & 5 & 10 & 240 \\ \hline
		$\mathcal{I}_{tst}$ & 101 & $\leq$50 & 101 & $\leq$1200 \\ \hline
	\end{tabular}
	\label{tab:caltech101_incl_settings}
	\vspace{-10pt}
\end{table}

In this experiment, the following four models are evaluated on the test set $\mathcal{I}_{tst}$ and the classification results are tabulated in Table \ref{tab:cltch101_inclusive}.
\begin{itemize}
	\setlength{\itemsep}{0pt}
	\item ResNet18\_C: ResNet18 network fine-tuned with the classical training approach on the training set $\mathcal{I}_{trn\_c}$.
	\item ResNet18\_F: ResNet18 network fine-tuned with the classical training approach on the training set $\mathcal{I}_{trn}$.
	\item ResNet18\_C-Full: ResNet18 network fine-tuned with the proposed approach on $\mathcal{I}_{trn\_c}$ and $\mathcal{I}_{op}$, where $\mathcal{I}_{op}$ includes all of the occlusion patterns.
	\item ResNet18\_F-Full: ResNet18 network fine-tuned with the proposed approach on $\mathcal{I}_{trn}$ and $\mathcal{I}_{op}$, where all occlusion patterns are included in $\mathcal{I}_{trn}$ and $\mathcal{I}_{op}$.	
	\vspace{-5pt}
\end{itemize} 

\begin{table*}
	\renewcommand{\arraystretch}{0.9}	
	\caption{Comparison of classification accuracy (\%) for the occlusion-inclusive training set on the Caltech101 dataset.}
	\centering
	\fontsize{8}{8}\selectfont
	\begin{tabular}{c|c|c|c|c|c|c}
		\hline
		Model&\multirow{2}{*}{\begin{tabular}[c]{@{}c@{}}Clean\\images\end{tabular}} &\multicolumn{2}{c|}{$20\%$ occlusion images} & \multicolumn{2}{c|}{$10\%$ occlusion images} & \multirow{2}{*}{\begin{tabular}[c]{@{}c@{}}Avg. over\\ occlusion\end{tabular}}  \\
		\cline{3-6}
		&  & center & random & center &random & \\
		\hline
		ResNet18\_F&90.05 & 42.06 & 48.51 &64.4 & 72.59& 58.72 \\	
		\hline
		ResNet18\_C&91.85 & 21.52 & 27.89 &46.41 & 57.47& 40.5 \\
		\hline
		ResNet18\_C-Full&91.61  & 44.24 & 54.74 &65.69& 75.35& 62.52 \\		
		\hline
		ResNet18\_F-Full&90.15 & 47.99 & 55.78 &69.26 & 76.14& 64.12 \\	
		\hline
	\end{tabular}
	\label{tab:cltch101_inclusive}
	\vspace{-5pt}
\end{table*}

\begin{figure}[h]
	\centering
	\includegraphics[width=0.5\linewidth]{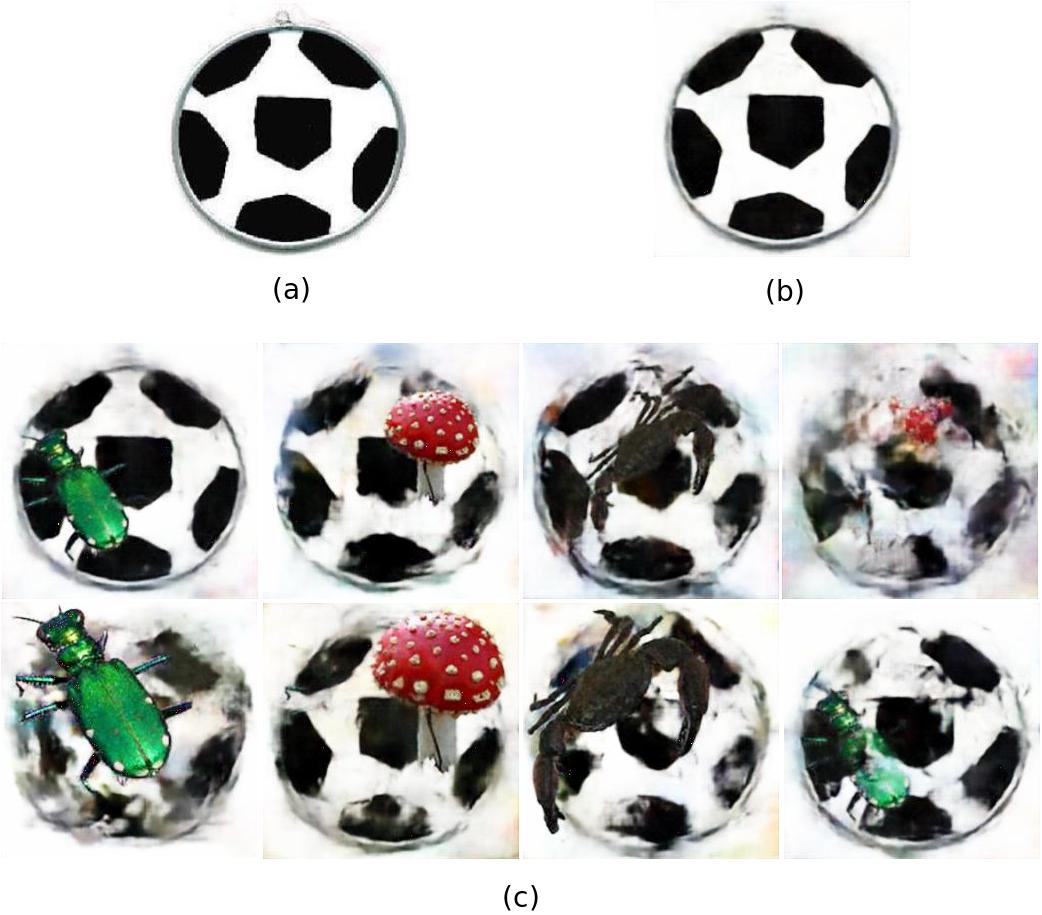}
	\vspace{-10pt}
	\caption{Visualization of the images reconstructed from the pseudo-DFVs. (a) the original image of $\mathbf{v}_i$; (b) the image reconstructed from $\mathbf{v}_i$; (c) the example images reconstructed from pseudo-DFVs $\widetilde{\mathbf{v}}_{ik}$'s. 
		\vspace{-10pt}
	}
	\label{fig:occ-free_decoding}	
\end{figure}
First, we provide a visual understanding of the pseudo-DFV by using a DFV decoder to project the pseudo-DFVs onto the image space. 
The DFV decoder is modified from the network of the generator of the deep convolutional generative adversarial networks (DCGAN)~\cite{radford2015unsupervised} by changing the input layer to $512$-D, adding two transposed convolutional layers to enlarge the output image from $64\times 64$ to $256\times 256$, changing the number of channels of each layer to adapt to the increased depth of the network, and employing a convolution layer as the output layer to improve the visualization quality.
The DFV decoder is trained on a synthetic dataset generated from the Caltech-101 dataset by employing the ResNet18\_C model as the DFV extractor (see Section S.II of the supplementary document for the structure and training detail of the DFV decoder).

The ResNet18\_C model is employed to extract the DFVs and the pseudo-DFVs are generated according to equation~(\ref{eq:psuedo_dfv}) by using the DVs extracted from $\mathcal{I}_{op}$ and setting $\beta=1$.
Some example images reconstructed from the pseudo-DFVs are shown in Fig.~\ref{fig:occ-free_decoding}.
From Fig.~\ref{fig:occ-free_decoding}, we can easily observe that the reconstructed image can be very similar to the synthetic occluded images (the first three columns of Fig.~\ref{fig:occ-free_decoding}(c)).
Although the occluders in some recovered images are blurred (the last column of Fig.~\ref{fig:occ-free_decoding}(c)), these reconstructed images still keep the image features of the "soccer\_ball" class.
In the experiment, we have not yet observed any images reconstructed from the pseudo-DFVs being incorrectly classified by human eyes. 
This also manifests the second observation obtained from Fig.~\ref{fig:pseudoDFVs}.

Then, we examine the impact of the occluded training images on the classification of clean images.
Apparently, with the occlusion-inclusive training set, the classical training techniques undermine the classification accuracy for the clean images due to the inconsistent statistics between the training set and the clean subset of the test images $\mathcal{I}_{tst\_c}$, \ie, the training set contains occluded images while $\mathcal{I}_{tst\_c}$ does not.
To ensure optimal average performance on the training set, the model trained with the classical training techniques has to make a compromise between the clean and the occluded training images.
The compromised model, however, is not optimal for the classification of clean images and thus the performance on $\mathcal{I}_{tst\_c}$ is degraded.

The results for the evaluations on the clean images in Table \ref{tab:cltch101_inclusive} manifest the degradation caused by the occluded training images (ResNet18\_F vs. ResNet18\_C and   ResNet18\_F-Full vs. ResNet18\_C-Full have $1.8\%$ and  $1.46\%$ decrease, respectively).
However, owing to the more compact representation induced by the back-propagation of the pseudo-DFVs (see section \ref{sec:correlation_dv}), the model trained with the proposed approach achieves smaller loss (ResNet18\_F-Full vs. ResNet18\_F achieves $0.1\%$ increase).

Finally, we compare the performances of the classical training approach and the proposed approach for the classification of occluded images.
From Table \ref{tab:cltch101_inclusive}, we can observe that the proposed approach with the training set either including or excluding the occluded images significantly improves the classification results (ResNet18\_F-Full vs. ResNet18\_F and ResNet18\_C-Full vs. ResNet18\_F achieve $5.4\%$ and $3.8\%$ increases, respectively).
In addition, the occluded images in the training set bring additional improvement for the proposed approach (ResNet18\_F-Full vs.ResNet18\_C-Full achieves $1.6\%$ increase).
These results confirm the analysis for the influence of the augmented DFVs in section \ref{sec:correlation_dv}. 

\subsection{AR Dataset}
The experiment in this section is designed to evaluate the proposed approach on a real occluded face dataset.
Each subject in the AR dataset has 26 images recorded in two different sessions (each with 13 images) separated by two weeks.
In each session, there are 7 images without occlusion, 3 images with sunglasses, and 3 images with scarves.
The set of clean training images $\mathcal{I}_{trn\_c}$ contains the images without occlusion in the first session (a total of 700 images).
The images without occlusion and with sunglasses or scarf in the first session from randomly selected 5 males and 5 females are taken as $\mathcal{I}_{op\_c}$ and $\mathcal{I}_{op\_o}$, respectively. 
Each image in $\mathcal{I}_{op\_c}$ is paired with each image of the same subject in $\mathcal{I}_{op\_o}$ to form the set of clean and occluded image pairs.
So, we obtain a total of 420 pairs of clean and occluded images.

The images with sunglasses or scarves in the first session from the rest 90 subjects and in the second session from all of the 100 subjects are used as the test images (a total of 1140 images).
To make fair comparisons with previous works, we evaluated our fine-tuned model on the test images in four different scenarios: 1) neutral faces with sunglasses,  2) all of the faces with
sunglasses, 3) neutral faces with scarves, and 4) all of the faces with scarves.

A publicly available pre-trained LResNet50E-IR network \cite{DBLP:journals/corr/abs-1801-07698}, which is a refined ResNet50 network, is adopted as the base CNN network for fine-tuning.
The LResNet50E-IR network is pre-trained on the MS-Celeb-1M dataset\cite{guo2016ms}.
To match the input size of the LResNet50E-IR network, all images are resized to $112\times 112$.
The following two models are evaluated in the above four scenarios.
\begin{itemize}
	\setlength{\itemsep}{0pt}
	\item LResNet50E-IR\_F: LResNet50E-IR network fine-tuned with the classical training approach on the training set $\mathcal{I}_{trn}=\mathcal{I}_{trn\_c}\cup \mathcal{I}_{op\_o}$.
	\item LResNet50E-IR\_C-Full: LResNet50E-IR network fine-tuned with the proposed approach on $\mathcal{I}_{trn\_c}$ and $\mathcal{I}_{op}$.
\end{itemize}

\begin{table}[h]	
	\renewcommand{\arraystretch}{0.9}
	\caption{
		Occluded-image examples and recognition accuracy (\%) on the AR dataset
	}
	\centering
	\fontsize{8}{8}\selectfont
	\includegraphics[width=0.5\linewidth]{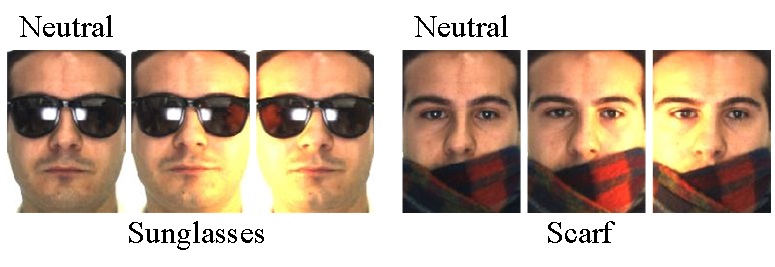}\\
	\begin{tabular}{l|c|c|c|c}
		\hline
		\multirow{2}{*}{} & \multicolumn{2}{c|}{Sunglasses}                                                         & \multicolumn{2}{c}{Scarf}                                                              \\ \cline{2-5} 
		& neutral               & all & neutral               & all \\ \hline
		SRC\cite{wright2009robust}& 87.0                  & -                                                               & 59.5                  & -                                                               \\ \hline
		GRRC\_$\text{L}_1$\cite{yang2013gabor}& 93.0                  & -                                                               & 79.0                  & -                                                               \\ \hline
		RNR\cite{qian2015robust} & -                     & 90.0                                                            & -                     & 86.2                                                             \\ \hline
		RDLRR\cite{gao2017learning} & -                     & 92.0                                                            & -                     & 91.0                                                            \\ \hline
		NMR\cite{yang2017nuclear}& -                     & 96.9                                                            & -                     & 73.5                                                            \\ \hline
		GD-HASLR\cite{wu2018occluded}& - & 93.2                                                            & - & 83.3                                                            \\ \hline
		MaskNet\cite{wan2017occlusion} &            -          & 90.9                                                            &  -                    & 96.7                                                            \\ \hline
		RPSM\cite{weng2016robust} & -                     & 96.0                                                            & -                     & 97.7                                                            \\ \hline
		PDSN\cite{Song2019OcclusionRF} & -                     & 99.7                                                            & -                     & 100                                                           \\ \hline
		LResNet50E-IR\_F             & 73.4                  & 72.1                                                            & 97.5                  & 96.9                                                            \\ \hline
		LResNet50E-IR\_C-Full      & 99.5                  & 98.0                                                            & 99.0                  & 99.0                                                            \\ \hline
	\end{tabular}
	\label{tab:AR_IR50_inclusive}
	\vspace{-10pt}
\end{table}
The face recognition results are tabulated in Table~\ref{tab:AR_IR50_inclusive}.
The results of other methods are obtained from their respective papers.
From Table~\ref{tab:AR_IR50_inclusive}, we can see that the LResNet50E-IR\_C-Full model achieves superior performance over the other approaches except for PSDN~\cite{Song2019OcclusionRF}, while the LResNet50E-IR\_F model shows the worst performance on the images with sunglasses.
Since in the training procedure these two models exploit the information of the same set of images but with different approaches, the good performance of the LResNet50E-IR\_C-Full model is credited to the proposed DFV augmentation approach.
Because the images of ten subjects with sunglasses or scarves in the first session are used in training, we verified these two models on the test images of the remaining 90 subjects.
Almost the same results (within $\pm0.3\%$) are obtained. 

Compared to PDSN, LResNet50E-IR\_C-Full shows a slightly worse performance.
However, it should be noted that PDSN requires an accurate occlusion detector to provide the shape and location of the occluder, while LResNet50E-IR\_C-Full does not.

\vspace{-10pt}
\subsection{ILSVRC2012}\label{sec:ILSVRC2012}
\vspace{-5pt}
The experiments in this section are designed to evaluate the proposed approach on the ILSVRC2012 dataset.

\begin{figure}[h]
	\centering
	\subfloat[]{
		\begin{minipage}[c][]{
				0.75\textwidth}
			\centering
			\includegraphics[width=1\textwidth]{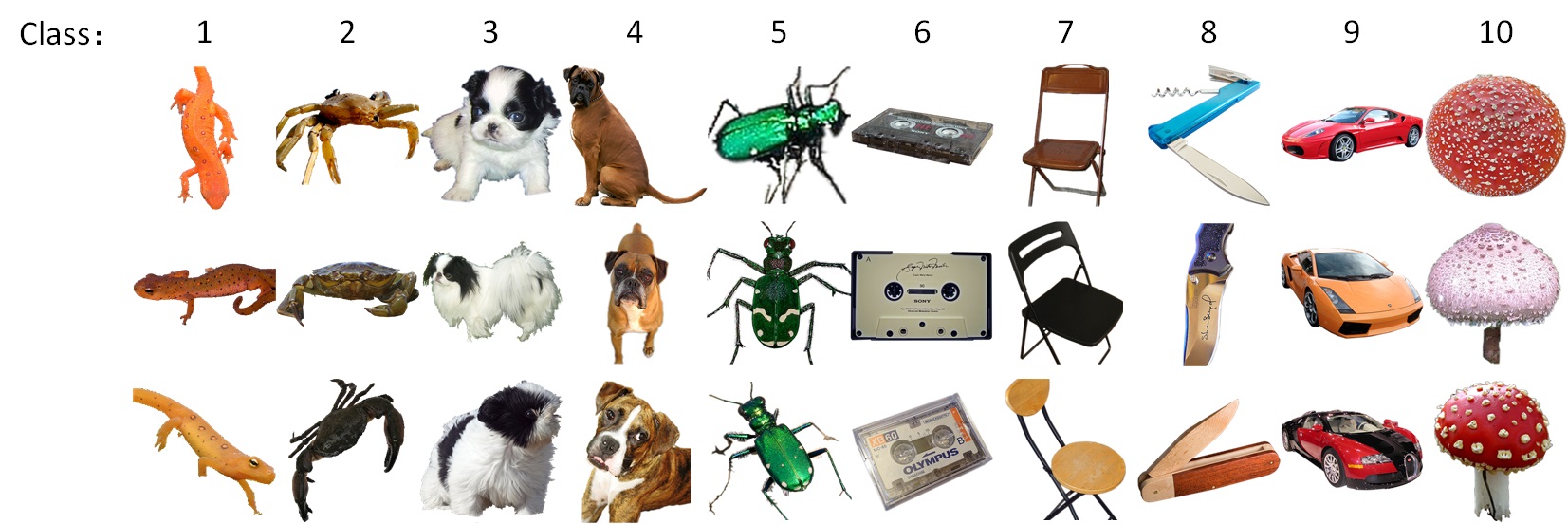}
		\end{minipage}
	}\\ \vspace{-10pt}
	\subfloat[]{
		\begin{minipage}[c][]{
				0.8\textwidth}
			\centering
			\includegraphics[width=1\textwidth]{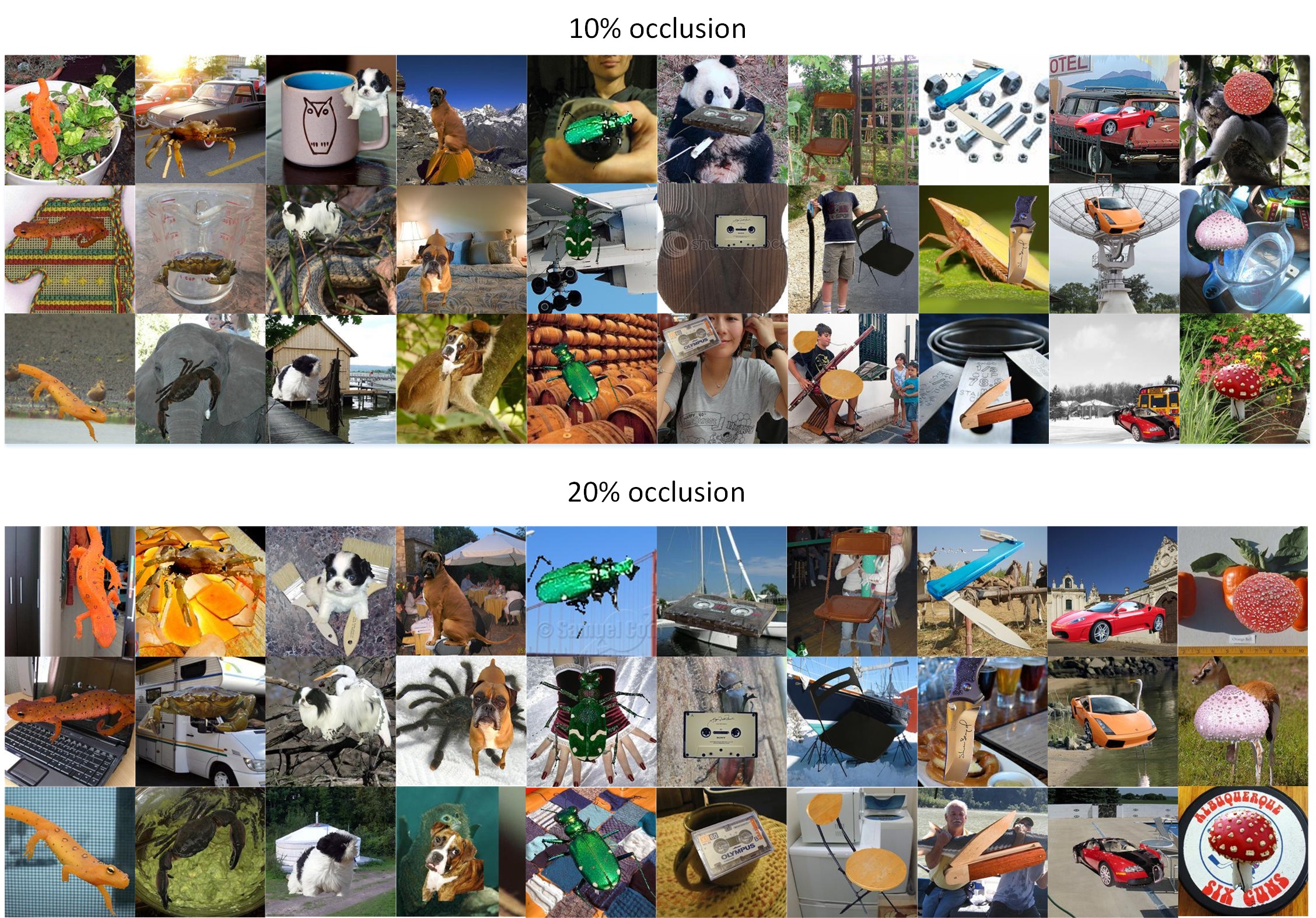}
		\end{minipage}
	}
	\caption{(a) Occluders used in the experiments and (b) examples of synthetic occluded test images for the experiments on the ILSVRC2012 dataset. 
		\vspace{-10pt}
	}
	\label{fig:occluder_imagenet}	
\end{figure}
In the experiments, the original $1000$ image classes are divided into two subsets: evaluation subset and extra subset.
The evaluation subset composed of $900$ image classes randomly drawn from the original $1000$ image classes is primarily used to construct the training set and test set. 
The extra subset consisting of the remaining $100$ image classes is primarily used to extract occluders and construct the set of clean and occluded image pairs for the experiment of the occlusion-exclusive training set.

240 occlusion patterns in total are employed to synthesize the occluded images.
$30$ occluders manually segmented from the images drawn from $10$ image classes, each with $3$ images, of the extra subset are shown in Fig.~\ref{fig:occluder_imagenet}(a).
Each occluder has two occlusion ratios ($10\%$ and $20\%$) and four corruption positions (a center position and three random positions).
Three random positions are independently generated for each occluder.
Some examples of the occluded test images for each occluder at two distinct occlusion ratios ($10\%$ and $20\%$) are shown in Fig.~\ref{fig:occluder_imagenet}(b). 
To save space, only one occlusion position is shown for each occluder at each specific occlusion ratio.

The original ILSVRC2012 validation images belonging to the evaluation subset are selected as the clean test images for all of the following experiments.
The same set of occluded test images is adopted for the experiments in Section \ref{sec:ILSVRC2012_exclusive}, \ref{sec:ILSVRC2012_inclusive}, and \ref{sec:ILSVRC2012_VGG}, where the occluded test images are synthesized by contaminating the clean test images with the 240 occlusion patterns.  

The ResNet50 network~\cite{he2016deep} pre-trained on the original ILSVRC2012 dataset is adopted as the base CNN for fine-tuning.
The hyperparameter $\alpha$ of the proposed approach is set to $0.9$.
In the experiments, all of the ResNet50 layers with the number of channels$\geq 512$ are included in fine-tuning and the data augmentation of random-resized-crop \cite{szegedy2015going} is adopted in the training of all the models.
The names of the models used for evaluation follow the naming protocol in Section \ref{sec:caltech101}.

\vspace{-8pt}
\subsubsection{Occlusion-exclusive training set} \label{sec:ILSVRC2012_exclusive}
In this experiment, the images of the ILSVRC2012 training set in the evaluation subset are employed as the training set $\mathcal{I}_{trn}$.
The set of clean and occluded image pairs $\mathcal{I}_{op}$ includes $50$ image classes randomly drawn from the extra subset except those used to extract the occluders.
The clean images of the clean and occluded image pairs, $5$ images per image class, are randomly drawn from the images of the $50$ image classes and the occluded images of the clean and occluded image pairs are synthesized by contaminating the clean images with all of the occlusion patterns.
Eventually, we obtain $250$ clean and occluded image pairs for each occlusion pattern.

Two models, ResNet50 and ResNet50-Full, fine-tuned with the classical training approach and the proposed approach, respectively, are adopted for evaluation.
The evaluation results are tabulated in Table \ref{tab:Imagenet_occlusion_exclusive}. It is evident that the proposed approach significantly improves the classification accuracy for the occluded images ($11.21\%$ increase on average and $13.94\%$ increase for randomly positioned $20\%$ occlusion).
\begin{table}
	\renewcommand{\arraystretch}{0.9}	
	\caption{Comparison of classification accuracy (\%) for the occlusion-exclusive training set on the ILSVRC2012 dataset.}
	\centering
	\fontsize{8}{8}\selectfont
	\begin{tabular}{c|c|c|c|c|c|c}
		\hline
		Model&\multirow{2}{*}{\begin{tabular}[c]{@{}c@{}}Clean\\images\end{tabular}} &\multicolumn{2}{c|}{$20\%$ occlusion images} & \multicolumn{2}{c|}{$10\%$ occlusion images} & \multirow{2}{*}{\begin{tabular}[c]{@{}c@{}}Avg. over\\ occlusion\end{tabular}}  \\
		\cline{3-6}
		&  & center & random & center &random & \\
		\hline
		ResNet50&74.65 & 26.99 & 30.77 &46.08 & 51.48& 39.98 \\	
		\hline
		ResNet50-Full&74.74 & 40.57 & 44.71 &54.92 & 59.98& 51.19 $(\uparrow 11.21)$ \\
		\hline
	\end{tabular}
	\label{tab:Imagenet_occlusion_exclusive}
	\vspace{-10pt}
\end{table}

\vspace{-5pt}
\subsubsection{Occlusion-inclusive training set}\label{sec:ILSVRC2012_inclusive}
In this experiment, the training set $\mathcal{I}_{trn}$ contains both clean images and occluded images.
$\mathcal{I}_{trn\_c}$ is the same as the training set $\mathcal{I}_{trn}$  used in Section \ref{sec:ILSVRC2012_exclusive}.
$\mathcal{I}_{op}$ is composed of $10$ image classes each with $5$ clean images.
The clean images of $\mathcal{I}_{op}$ are randomly drawn from
$\mathcal{I}_{trn\_c}$.
$\mathcal{I}_{op\_o}$ is adopted as $\mathcal{I}_{trn\_o}$.

\begin{table}[h]
	\renewcommand{\arraystretch}{0.9}
	\vspace{-10pt}	
	\caption{Comparison of classification accuracy (\%) for the occlusion-inclusive training set on the ILSVRC2012 dataset.}
	\centering
	\fontsize{8}{8}\selectfont
	\begin{tabular}{c|c|c|c|c|c|c}
		\hline
		Model&\multirow{2}{*}{\begin{tabular}[c]{@{}c@{}}Clean\\images\end{tabular}} &\multicolumn{2}{c|}{$20\%$ occlusion images} & \multicolumn{2}{c|}{$10\%$ occlusion images} & \multirow{2}{*}{\begin{tabular}[c]{@{}c@{}}Avg. over\\ occlusion\end{tabular}}  \\
		\cline{3-6}
		&  & center & random & center &random & \\
		\hline
		ResNet50\_F &74.39 & 34.54 & 38.17 &51.13 & 55.83& 45.96 \\	
		\hline
		 ResNet50\_C-Full&74.72 & 40.88 & 45.14 &55.26 & 60.31& 51.56 $(\uparrow 5.6)$ \\
		\hline
		 ResNet50\_F-Full&74.73  & 46.1 & 49.8 &58.19& 62.38& 55.1 $(\uparrow 9.14)$ \\
		\hline
	\end{tabular}
	\label{tab:Imagenet_occlusion_inclusive}
	\vspace{-10pt}
\end{table}
Three models, ResNet50\_F, ResNet50\_C-Full, and ResNet50\_F-Full, are evaluated and the results are reported in Table \ref{tab:Imagenet_occlusion_inclusive}.
It can be seen from the results that, even though the training set includes occluded images, the proposed approach significantly boost the classification accuracy for occluded images (ResNet50\_F-Full vs. ResNet50\_F: $9.14\%$ increase on average and $11.63\%$ increase for randomly positioned $20\%$ occlusion).
By comparing ResNet50 in Table \ref{tab:Imagenet_occlusion_exclusive} with ResNet50\_F in Table \ref{tab:Imagenet_occlusion_inclusive}, we observe that 
the occlusion-inclusive training set degrades the classification accuracy for the clean images for the classical training approach ($0.26\%$ decrease).
On the contrary, the proposed approach slightly improves the classification of the clean images (ResNet50\_F-Full in Table \ref{tab:Imagenet_occlusion_inclusive} vs. ResNet50 in Table \ref{tab:Imagenet_occlusion_exclusive}: $0.08\%$ increase).
This observation in combination with the observation in Section \ref{sec:caltech_occ_inc} indicates that the proposed approach has the advantage to alleviate the degradation caused by the occluded training images for the classification of the clean images.

In addition, similar classification accuracy achieved by ResNet50\_C-Full in Table~\ref{tab:Imagenet_occlusion_inclusive} and ResNet50-Full in Table~\ref{tab:Imagenet_occlusion_exclusive}, both of which have the same settings except that they use distinct clean images to construct $\mathcal{I}_{op}$, indicates that the proposed approach is not sensitive to the clean images employed to construct $\mathcal{I}_{op}$. 
\vspace{-10pt}
\subsubsection{Cross-pattern}
This experiment is designed to assess the cross-pattern performance of the proposed approach.
Here, the cross-pattern refers to the setting that the occlusion patterns in $\mathcal{I}_{tst\_o}$ (the occluded image subset of $\mathcal{I}_{tst}$) are distinct from those in $\mathcal{I}_{op\_o}$ (the occluded image subset of $\mathcal{I}_{op}$).

In the experiment, the $\mathcal{I}_{trn\_c}$, $\mathcal{I}_{tst\_c}$ and $\mathcal{I}_{op\_c}$ used in Section \ref{sec:ILSVRC2012_inclusive} are adopted as $\mathcal{I}_{trn\_c}$, $\mathcal{I}_{tst\_c}$ and $\mathcal{I}_{op\_c}$, respectively.
To ensure non-overlapping between the occlusion patterns in $\mathcal{I}_{tst\_o}$ and $\mathcal{I}_{op\_o}$, the first row of Fig.~\ref{fig:occluder_imagenet}(a) are employed to synthesize the occluded images in $\mathcal{I}_{op\_o}$ and the second and third row of Fig.~\ref{fig:occluder_imagenet}(a) to produce the occluded images in $\mathcal{I}_{tst\_o}$.
The training set is set to occlusion-inclusion, \ie, $\mathcal{I}_{trn} = \mathcal{I}_{trn\_c}\cup \mathcal{I}_{op\_o}$.
Two models, ResNet50\_F and ResNet50\_F-Full, which are fine-tuned on respective $\mathcal{I}_{trn}$ and $\mathcal{I}_{trn}$ and $\mathcal{I}_{op}$, are evaluated on $\mathcal{I}_{tst}$.

\begin{table*}
	\renewcommand{\arraystretch}{0.9}	
	\caption{Comparison of classification accuracy (\%) for the cross-pattern setting on the ILSVRC2012 dataset.}
	\centering
	\fontsize{8}{8}\selectfont
	\begin{tabular}{c|c|c|c|c|c|c}
		\hline
		Model&\multirow{2}{*}{\begin{tabular}[c]{@{}c@{}}Clean\\images\end{tabular}} &\multicolumn{2}{c|}{$20\%$ occlusion images} & \multicolumn{2}{c|}{$10\%$ occlusion images} & \multirow{2}{*}{\begin{tabular}[c]{@{}c@{}}Avg. over\\ occlusion\end{tabular}}  \\
		\cline{3-6}
		&  & center & random & center &random & \\
		\hline
		ResNet50\_F&74.54 & 30.45 & 34.65 &49.36 & 53.50& 43.03 \\	
		\hline
		ResNet50\_F-Full&74.62 & 34.90 & 39.07 &52.57 & 56.83& 46.90 $(\uparrow 3.87)$ \\
		\hline
	\end{tabular}
	\label{tab:Imagenet_occ_inc_intra_pattern_cross}
	\vspace{-10pt}
\end{table*}

The experimental results are tabulated in Table \ref{tab:Imagenet_occ_inc_intra_pattern_cross}, from which we can observe that the proposed approach can improve the classification accuracy ($3.87\%$ increase averaged on occlusion patterns) for the images occluded with the occlusion patches unseen in the training phase. This property will greatly benefit the practical applications.

We can also note that the occlusion patches in each column of Fig.~\mbox{\ref{fig:occluder_imagenet}}(a) belong to the same category. The experimental results imply that this kind of cross-pattern DVs locate, to a certain extent, close to each other on a low-dimensional manifold and can be linearly approximated by each other to generate the pseudo-DFVs. However, for other types of cross-pattern DVs, further investigation needs to be conducted.

\vspace{-10pt}
\subsubsection{Additional networks} \label{sec:ILSVRC2012_VGG}
In this experiment, the VGG16 networks \cite{simonyan2014very} are adopted.
The experiment settings are the same as those in Section \ref{sec:ILSVRC2012_inclusive} for the occlusion-inclusive training set.

Four models, VGG16\_C, VGG16\_F, VGG16\_C-Full, and VGG16\_F-Full, named after Section \ref{sec:caltech_occ_inc} except for distinct netwoks, are assessed for comparison.
The classification results for each model and the improvements, VGG16\_C-Full \wrt VGG16\_C and VGG16\_F-Full \wrt VGG16\_F, are tabulated in Table~\ref{tab:ImageNetVGG16_inclusive}.
The results demonstrate that the proposed approach is a generic technique to improve the classification performance for occluded images.
\begin{table}[h]
	\vspace{-10pt}
	\renewcommand{\arraystretch}{0.9}	
	\caption{Comparison of classification accuracy (\%) for additional networks on the ILSVRC2012 dataset.}
	\centering
	\fontsize{8}{8}\selectfont
	\begin{tabular}{l|c|c}
		\hline
		Model      & Clean images            &  \begin{tabular}[c]{@{}c@{}}Occluded images \\ Avg. over occlusion \end{tabular}       \\ \hline
		VGG16\_C      &  70.25                  & 34.8     \\ \hline
		VGG16\_C-Full   & 70.54 & 39.48 $(\uparrow 4.68)$  \\ \hline \hline
		VGG16\_F   & 70.03                   & 46.72    \\ \hline
		VGG16\_F-Full & 70.07  & 49.44 $(\uparrow 2.72)$ \\ \hline
	\end{tabular}
	\label{tab:ImageNetVGG16_inclusive}
	\vspace{-10pt}
\end{table}

It should also be noted that, in all of the above experiments, all models are trained along with the data augmentation and the proposed approach has similar convergence speed as the classical training approach.
Therefore, the proposed approach is compatible with the data augmentation and does not induce much extra computational complexity in fine-tuning.
\vspace{-10pt}
\section{Conclusion}\label{sec:conclusion}
\vspace{-5pt}
In this paper, we have proposed a novel deep feature augmentation approach for fine-tuning.
The proposed approach significantly boosts the classification accuracy for the occluded images.
For the classification of clean images, the models fine-tuned with the proposed approach achieve comparable classification accuracy to those fine-tuned with the classical training approach. For the training set containing occluded training images, compared to the classical training approach, the proposed approach effectively alleviates the performance degradation caused by the occluded training images in the classification of clean images. 
In addition, we have synthesized a large-scale occluded image dataset from ILSVRC2012 dataset~\cite{ILSVRC15}. 

Although the proposed approach exhibits promising performance on the classification of occluded images, further investigations need to be done in practical applications, such as a comprehensive understanding of the deep feature augmentation for the cross-pattern, to promote the performance for the cross-pattern, and exploring the extra image pairs formed by flexible approaches, \ie, the occluded image belongs to the same image class but not the corrupted version of the clean image.
The proposed approach can also be applied to improve the performance of other computer vision tasks, such as object detection and semantic segmentation, since all these algorithms are based on CNNs.

\vspace{-10pt}
\section*{Acknowledgment}
\vspace{-5pt}
This work was partly supported by Shanghai Agriculture Applied Technology Development Program, China, under Grant G20180306.

\vspace{-5pt}
\bibliographystyle{elsarticle-num}
\bibliography{arXiv_dfv_aug}

\begin{thebibliography}{10}
\expandafter\ifx\csname url\endcsname\relax
  \def\url#1{\texttt{#1}}\fi
\expandafter\ifx\csname urlprefix\endcsname\relax\def\urlprefix{URL }\fi
\expandafter\ifx\csname href\endcsname\relax
  \def\href#1#2{#2} \def\path#1{#1}\fi

\bibitem{liu1990partial}
H.-C. Liu, M.~D. Srinath, Partial shape classification using contour matching
  in distance transformation, IEEE Transactions on Pattern Analysis and Machine
  Intelligence 12~(11) (1990) 1072--1079.

\bibitem{zhang2003object}
J.~Zhang, X.~Zhang, H.~Krim, G.~G. Walter, Object representation and
  recognition in shape spaces, Pattern Recognition 36~(5) (2003) 1143--1154.

\bibitem{huang1997object}
C.-Y. Huang, O.~I. Camps, T.~Kanungo, Object recognition using appearance-based
  parts and relations, in: Proceedings of IEEE Computer Society Conference on
  Computer Vision and Pattern Recognition, 1997, pp. 877--883.

\bibitem{ohba1997detectability}
K.~Ohba, K.~Ikeuchi, Detectability, uniqueness, and reliability of eigen
  windows for stable verification of partially occluded objects, IEEE
  Transactions on Pattern Analysis and Machine Intelligence 19~(9) (1997)
  1043--1047.

\bibitem{yang2013gabor}
M.~Yang, L.~Zhang, S.~C. Shiu, D.~Zhang, Gabor feature based robust
  representation and classification for face recognition with gabor occlusion
  dictionary, Pattern Recognition 46~(7) (2013) 1865--1878.

\bibitem{ou2014robust}
W.~Ou, X.~You, D.~Tao, P.~Zhang, Y.~Tang, Z.~Zhu, Robust face recognition via
  occlusion dictionary learning, Pattern Recognition 47~(4) (2014) 1559--1572.

\bibitem{yu2017discriminative}
Y.-F. Yu, D.-Q. Dai, C.-X. Ren, K.-K. Huang, Discriminative multi-scale sparse
  coding for single-sample face recognition with occlusion, Pattern Recognition
  66 (2017) 302--312.

\bibitem{zhou2009face}
Z.~Zhou, A.~Wagner, H.~Mobahi, J.~Wright, Y.~Ma, Face recognition with
  contiguous occlusion using markov random fields, in: Proceedings of the IEEE
  International Conference on Computer Vision, 2009, pp. 1050--1057.

\bibitem{jia2012robust}
K.~Jia, T.-H. Chan, Y.~Ma, Robust and practical face recognition via structured
  sparsity, in: Proceedings of the European conference on computer vision,
  2012, pp. 331--344.

\bibitem{li2013structured}
X.-X. Li, D.-Q. Dai, X.-F. Zhang, C.-X. Ren, Structured sparse error coding for
  face recognition with occlusion, IEEE transactions on image processing 22~(5)
  (2013) 1889--1900.

\bibitem{yang2017nuclear}
J.~Yang, L.~Luo, J.~Qian, Y.~Tai, F.~Zhang, Y.~Xu, Nuclear norm based matrix
  regression with applications to face recognition with occlusion and
  illumination changes, IEEE transactions on pattern analysis and machine
  intelligence 39~(1) (2017) 156--171.

\bibitem{deng2012extended}
W.~Deng, J.~Hu, J.~Guo, Extended {SRC}: {Undersampled} face recognition via
  intraclass variant dictionary, IEEE Transactions on Pattern Analysis and
  Machine Intelligence 34~(9) (2012) 1864--1870.

\bibitem{wu2018occluded}
C.~Y. Wu, J.~J. Ding, Occluded face recognition using low-rank regression with
  generalized gradient direction, Pattern Recognition 80 (2018) 256--268.

\bibitem{simonyan2014very}
K.~Simonyan, A.~Zisserman, Very deep convolutional networks for large-scale
  image recognition, arXiv preprint arXiv:1409.1556 (2014).

\bibitem{he2016deep}
K.~He, X.~Zhang, S.~Ren, J.~Sun, Deep residual learning for image recognition,
  in: Proceedings of the {IEEE} Conference on Computer Vision and Pattern
  Recognition, 2016, pp. 770--778.

\bibitem{szegedy2017inception}
C.~Szegedy, S.~Ioffe, V.~Vanhoucke, A.~A. Alemi, Inception-v4, inception-resnet
  and the impact of residual connections on learning., in: AAAI, Vol.~4, 2017,
  p.~12.

\bibitem{cen2019boosting}
F.~Cen, G.~Wang, Boosting occluded image classification via subspace
  decomposition-based estimation of deep features, {IEEE} Transactions on
  Cybernetics 50~(7) (2020) 3409--3422.

\bibitem{cen2019dictionary}
F.~Cen, G.~Wang, Dictionary representation of deep features for
  occlusion-robust face recognition, IEEE Access 7 (2019) 26595--26605.

\bibitem{ma2020mdfn}
W.~Ma, Y.~Wu, F.~Cen, G.~Wang, {MDFN}: Multi-scale deep feature learning
  network for object detection, Pattern Recognition 100 (2020) 107149.

\bibitem{ettinger1988large}
G.~J. Ettinger, Large hierarchical object recognition using libraries of
  parameterized model sub-parts, in: Proceedings of the {IEEE} Conference on
  Computer Vision and Pattern Recognition, 1988, pp. 32--41.

\bibitem{tsang1994classification}
P.~W.-M. Tsang, P.~C. Yuen, F.~Lam, Classification of partially occluded
  objects using 3-point matching and distance transformation, Pattern
  recognition 27~(1) (1994) 27--40.

\bibitem{li2001learning}
S.~Z. Li, X.~Hou, H.~Zhang, Q.~Cheng, Learning spatially localized, parts-based
  representation, in: Proceedings of the {IEEE} Computer Society Conference on
  Computer Vision and Pattern Recognition, 2001, pp. 207--212.

\bibitem{martinez2002recognizing}
A.~M. Mart{\'\i}nez, Recognizing imprecisely localized, partially occluded, and
  expression variant faces from a single sample per class, IEEE Transactions on
  Pattern Analysis and Machine Intelligence 24~(6) (2002) 748--763.

\bibitem{tan2005recognizing}
X.~Tan, S.~Chen, Z.-H. Zhou, F.~Zhang, Recognizing partially occluded,
  expression variant faces from single training image per person with som and
  soft k-nn ensemble, IEEE Transactions on Neural Networks 16~(4) (2005)
  875--886.

\bibitem{kim2005effective}
J.~Kim, J.~Choi, J.~Yi, M.~Turk, Effective representation using {ICA} for face
  recognition robust to local distortion and partial occlusion, IEEE
  Transactions on Pattern Analysis \& Machine Intelligence 27~(12) (2005)
  1977--1981.

\bibitem{wright2009robust}
J.~Wright, A.~Y. Yang, A.~Ganesh, S.~S. Sastry, Y.~Ma, Robust face recognition
  via sparse representation, IEEE Transactions on Pattern Analysis and Machine
  Intelligence 31~(2) (2009) 210--227.

\bibitem{gao2017learning}
G.~Gao, J.~Yang, X.-Y. Jing, F.~Shen, W.~Yang, D.~Yue, Learning robust and
  discriminative low-rank representations for face recognition with occlusion,
  Pattern Recognition 66 (2017) 129--143.

\bibitem{pathak2016context}
D.~Pathak, P.~Krahenbuhl, J.~Donahue, T.~Darrell, A.~A. Efros, Context
  encoders: Feature learning by inpainting, in: Proceedings of the IEEE
  Conference on Computer Vision and Pattern Recognition, 2016, pp. 2536--2544.

\bibitem{nguyen2017plug}
A.~Nguyen, J.~Clune, Y.~Bengio, A.~Dosovitskiy, J.~Yosinski, Plug \& play
  generative networks: Conditional iterative generation of images in latent
  space, in: Proceedings of the IEEE Conference on Computer Vision and Pattern
  Recognition, 2017, pp. 3510--3520.

\bibitem{yang2017high}
C.~Yang, X.~Lu, Z.~Lin, E.~Shechtman, O.~Wang, H.~Li, High-resolution image
  inpainting using multi-scale neural patch synthesis, in: Proceedings of the
  {IEEE} Conference on Computer Vision and Pattern Recognition, 2017, pp.
  4076--4084.

\bibitem{yu2018generative}
J.~Yu, Z.~Lin, J.~Yang, X.~Shen, X.~Lu, T.~S. Huang, Generative image
  inpainting with contextual attention, in: Proceedings of the IEEE Conference
  on Computer Vision and Pattern Recognition, 2018, pp. 5505--5514.

\bibitem{cheng2015robust}
L.~Cheng, J.~Wang, Y.~Gong, Q.~Hou, Robust deep auto-encoder for occluded face
  recognition, in: Proceedings of the 23rd ACM international conference on
  Multimedia, ACM, 2015, pp. 1099--1102.

\bibitem{zhao2018robust}
F.~Zhao, J.~Feng, J.~Zhao, W.~Yang, S.~Yan, Robust {LSTM}-autoencoders for face
  de-occlusion in the wild, IEEE Transactions on Image Processing 27~(2) (2018)
  778--790.

\bibitem{DBLP:journals/corr/abs-1711-04340}
A.~Antoniou, A.~J. Storkey, H.~Edwards, Data augmentation generative
  adversarial networks, CoRR abs/1711.04340 (2017).
\newblock \href {http://arxiv.org/abs/1711.04340} {\path{arXiv:1711.04340}}.

\bibitem{Song2019OcclusionRF}
L.~Song, D.~Gong, Z.~Li, C.~Liu, W.~Liu, Occlusion robust face recognition
  based on mask learning with pairwise differential {Siamese} network, 2019
  IEEE/CVF International Conference on Computer Vision (ICCV) (2019) 773--782.

\bibitem{juefei2016deepgender}
F.~Juefei-Xu, E.~Verma, P.~Goel, A.~Cherodian, M.~Savvides, {DeepGender:
  Occlusion and Low Resolution Robust Facial Gender Classification via
  Progressively Trained Convolutional Neural Network with Attention}, in:
  Proceedings of the IEEE Conference on Computer Vision and Pattern Recognition
  Workshops (CVPRW), IEEE, 2016, pp. 136--145.

\bibitem{fei2007learning}
L.~Fei-Fei, R.~Fergus, P.~Perona, Learning generative visual models from few
  training examples: An incremental bayesian approach tested on 101 object
  categories, Computer vision and Image understanding 106~(1) (2007) 59--70.

\bibitem{maaten2008visualizing}
L.~v.~d. Maaten, G.~Hinton, Visualizing data using t-{SNE}, Journal of Machine
  Learning Research 9~(Nov) (2008) 2579--2605.

\bibitem{martinez1998ar}
A.~M. Martinez, The {AR} face database, CVC Technical Report24 (1998).

\bibitem{ILSVRC15}
O.~Russakovsky, J.~Deng, H.~Su, J.~Krause, S.~Satheesh, S.~Ma, Z.~Huang,
  A.~Karpathy, A.~Khosla, M.~Bernstein, A.~C. Berg, L.~Fei-Fei, {ImageNet Large
  Scale Visual Recognition Challenge}, International Journal of Computer Vision
  115~(3) (2015) 211--252.

\bibitem{radford2015unsupervised}
A.~Radford, L.~Metz, S.~Chintala, Unsupervised representation learning with
  deep convolutional generative adversarial networks, arXiv preprint
  arXiv:1511.06434 (2015).

\bibitem{DBLP:journals/corr/abs-1801-07698}
J.~Deng, J.~Guo, S.~Zafeiriou, Arcface: Additive angular margin loss for deep
  face recognition, CoRR abs/1801.07698 (2018).
\newblock \href {http://arxiv.org/abs/1801.07698} {\path{arXiv:1801.07698}}.

\bibitem{guo2016ms}
Y.~Guo, L.~Zhang, Y.~Hu, X.~He, J.~Gao, {Ms-celeb-1m: A dataset and benchmark
  for large-scale face recognition}, in: European Conference on Computer
  Vision, Springer, 2016, pp. 87--102.

\bibitem{qian2015robust}
J.~Qian, L.~Luo, J.~Yang, F.~Zhang, Z.~Lin, Robust nuclear norm regularized
  regression for face recognition with occlusion, Pattern Recognition 48~(10)
  (2015) 3145--3159.

\bibitem{wan2017occlusion}
W.~Wan, J.~Chen, Occlusion robust face recognition based on mask learning, in:
  2017 IEEE International Conference on Image Processing (ICIP), IEEE, 2017,
  pp. 3795--3799.

\bibitem{weng2016robust}
R.~Weng, J.~Lu, Y.-P. Tan, Robust point set matching for partial face
  recognition, IEEE transactions on image processing 25~(3) (2016) 1163--1176.

\bibitem{szegedy2015going}
C.~Szegedy, W.~Liu, Y.~Jia, P.~Sermanet, S.~Reed, D.~Anguelov, D.~Erhan,
  V.~Vanhoucke, A.~Rabinovich, Going deeper with convolutions, in: Proceedings
  of the {IEEE} Conference on Computer Vision and Pattern Recognition, 2015,
  pp. 1--9.

\end{thebibliography}


\begin{thebibliography}{1}
\expandafter\ifx\csname url\endcsname\relax
  \def\url#1{\texttt{#1}}\fi
\expandafter\ifx\csname urlprefix\endcsname\relax\def\urlprefix{URL }\fi
\expandafter\ifx\csname href\endcsname\relax
  \def\href#1#2{#2} \def\path#1{#1}\fi

\bibitem{maaten2008visualizing}
L.~v.~d. Maaten, G.~Hinton, Visualizing data using t-{SNE}, Journal of Machine
  Learning Research 9~(Nov) (2008) 2579--2605.

\bibitem{radford2015unsupervised}
A.~Radford, L.~Metz, S.~Chintala, Unsupervised representation learning with
  deep convolutional generative adversarial networks, arXiv preprint
  arXiv:1511.06434 (2015).

\bibitem{han2018attribute}
K.~Han, J.~Guo, C.~Zhang, M.~Zhu, Attribute-aware attention model for
  fine-grained representation learning, in: 2018 ACM Multimedia Conference on
  Multimedia Conference, ACM, 2018, pp. 2040--2048.

\bibitem{WahCUB_200_2011}
C.~Wah, S.~Branson, P.~Welinder, P.~Perona, S.~Belongie, {The Caltech-UCSD
  Birds-200-2011 Dataset}, Tech. Rep. CNS-TR-2011-001, California Institute of
  Technology (2011).

\bibitem{iamhankai_a3m_codes}
\href{https://github.com/iamhankai/attribute-aware-attention}{[link]}.
\newline\urlprefix\url{https://github.com/iamhankai/attribute-aware-attention}

\end{thebibliography}

\vfill
\end{document}


\title{Deep Feature Augmentation for Occluded Image Classification\\--Supplementary Document}
\author{Feng~Cen,
	Xiaoyu~Zhao, Wuzhuang~Li, 
	Guanghui~Wang
	}
\maketitle

\section{Visualization of DFVs and DVs}
\begin{figure*}
	\centering
	\includegraphics[width=0.8\linewidth]{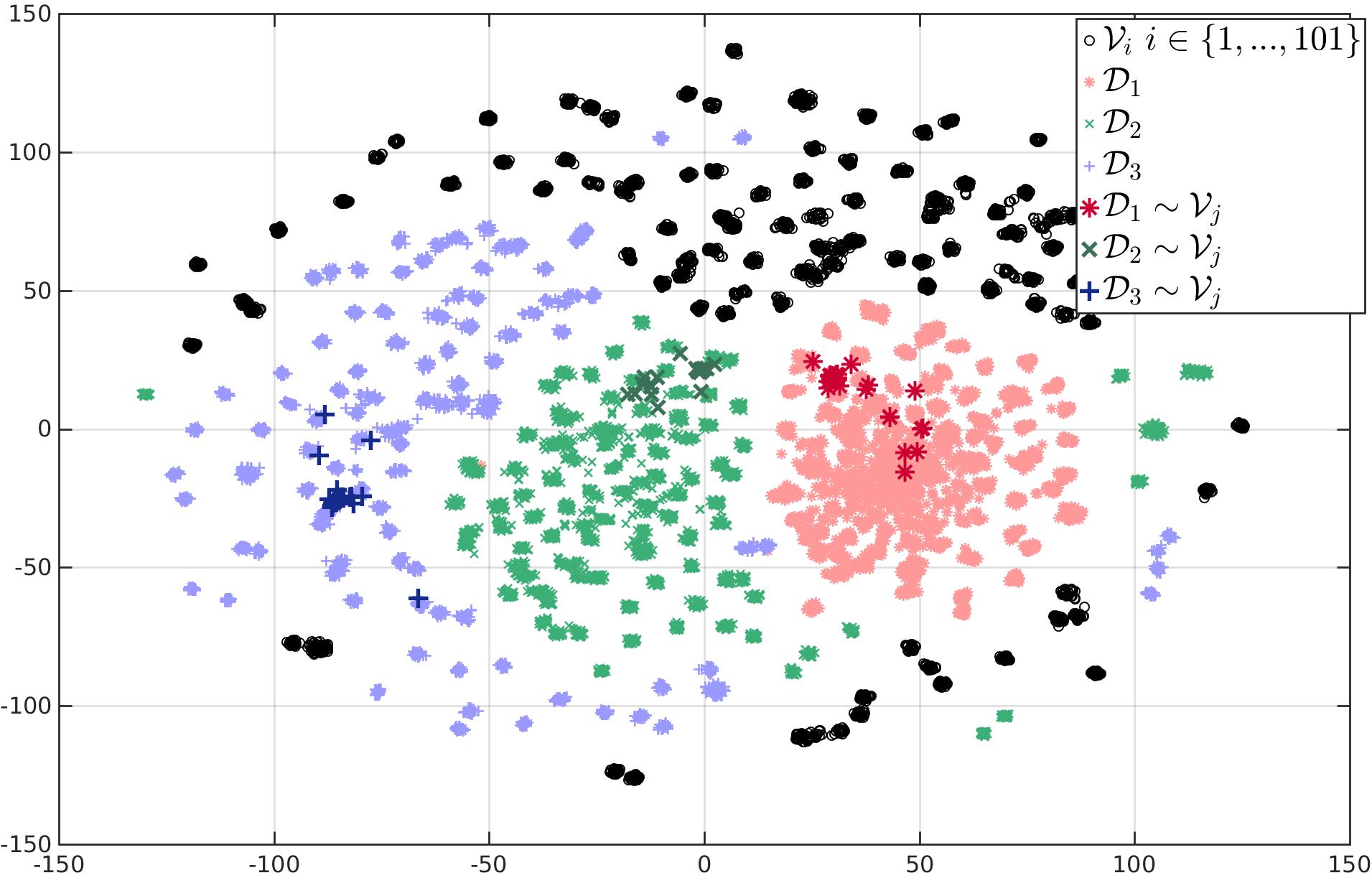}
	\caption{2-D illustration of DFVs and DVs for the caltech-101 dataset. The t-SNE algorithm~\mbox{\cite{maaten2008visualizing}} is employed to project the DFVs and DVs from $512$-D to $2$-D for visualization. $\mathcal{D}_k\sim \mathcal{V}_j$ denotes the set of the DVs associated with the $k$th occlusion pattern and generated from the images of the $j$th image class. The $j$th image class is a randomly selected image class. (better viewed in color) }
	\label{sfig:vis_dfv_dv}
\end{figure*}

In this section, we employ the t-SNE algorithm to visualize the DFVs and DVs generated from the 101 image classes of the Caltech-101 dataset.
The clean image subset, $\mathcal{I}_{trn\_c}$, of the training set used in Section~4.1.2, 30 images per image class, is adopted as the clean image set.
The occluded images are synthesized from the clean image set by using three $20\%$ center occlusion patterns shown in Fig.~5.
The ResNet18\_C model in Section~4.1.2 is adopted as the DFV extractor. 
In Fig.~\mbox{S\ref{sfig:vis_dfv_dv}}, the DFVs are extracted from the clean image set and the DVs are generated from the clean and occluded image pairs.

From Fig.~\mbox{S\ref{sfig:vis_dfv_dv}}, we can observe that the DVs associated with one occlusion pattern can easily be separated from the DVs associated with the other two occlusion patterns and the DFVs of the clean images.
This observation manifests the fact that the intra-pattern DVs are close to each other on a low-dimensional manifold in the deep feature space.

We can also observe that for each occlusion pattern, the DVs in $\mathcal{D}_k\sim \mathcal{V}_j$ do not locate in a small cluster.
This indicates that by employing a subset of the clean images in a subset of the image classes, we can generate the DVs covering most of the space associated with the occlusion patterns, and then, by using these DVs and the rest of the clean images in the dataset, we can generate the pseudo-DFVs covering most of the space occupied by the DFVs of the occluded images.

\section{Image recovering from pseudo-DFV}
In this section, we elaborate on the network structure and training procedure of the DFV decoder.
Additional examples of the reconstructed images are also provided in this section.    

\subsection{Network structure}
\begin{figure*}
	\centering
	\includegraphics[width=0.9\linewidth]{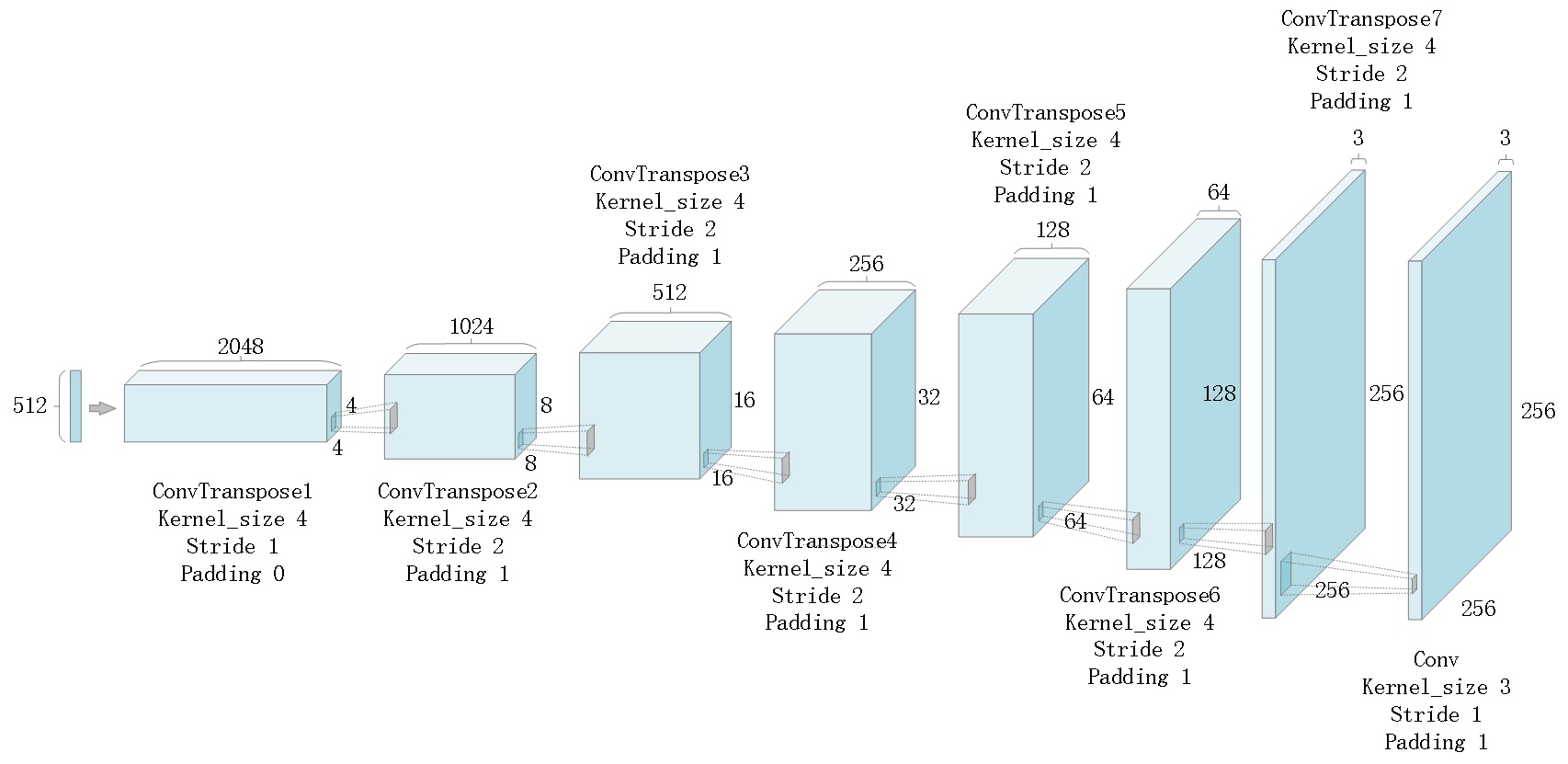}
	\caption{The network structure of the DFV decoder }
	\label{sfig:dfv_decoder_structure}
\end{figure*}

The network structure of the DFV decoder is shown in Fig.~S\ref{sfig:dfv_decoder_structure}.
The network is modified from the generator network of the DCGAN \cite{radford2015unsupervised}.
The modification includes:
1) the input layer is changed to $512$-D to match the size of the DFV;
2) two transposed convolutional layers are added to enlarge the output image from $64\times 64$ to $256\times 256$;
3) the number of channels of each layer is changed (as shown in Fig.~S\ref{sfig:dfv_decoder_structure}) to adapt to the increased depth of the network;
4) a convolution layer is employed as the output layer to improve the visualization quality.

\subsection{Training DFV decoder}
The DFV decoder is trained on a synthetic dataset containing both clean images and synthetic occluded images.
All of the images in the Caltech-101 dataset are adopted as the clean training images. 
The synthetic occluded images are generated by applying each occlusion pattern in Fig.~5 to each clean training image.

The DFVs of the training images are extracted by using the ResNet18\_C model in Section 4.1.2 and the training images are resized to $224\times 224$ to fit the input size of the ResNet18\_C model.
In DFV decoder training, the training images are resized to $256\times 256$ as the ground truth of the output and the DFVs of the training images are input to the DFV decoder for network training.
The mean square error (MSE) measured between the output of the DFV decoder and the ground truth image is adopted as the loss.

\subsection{Additional examples of recovered images}
\begin{figure}
	\centering
	\includegraphics[width=0.8\linewidth]{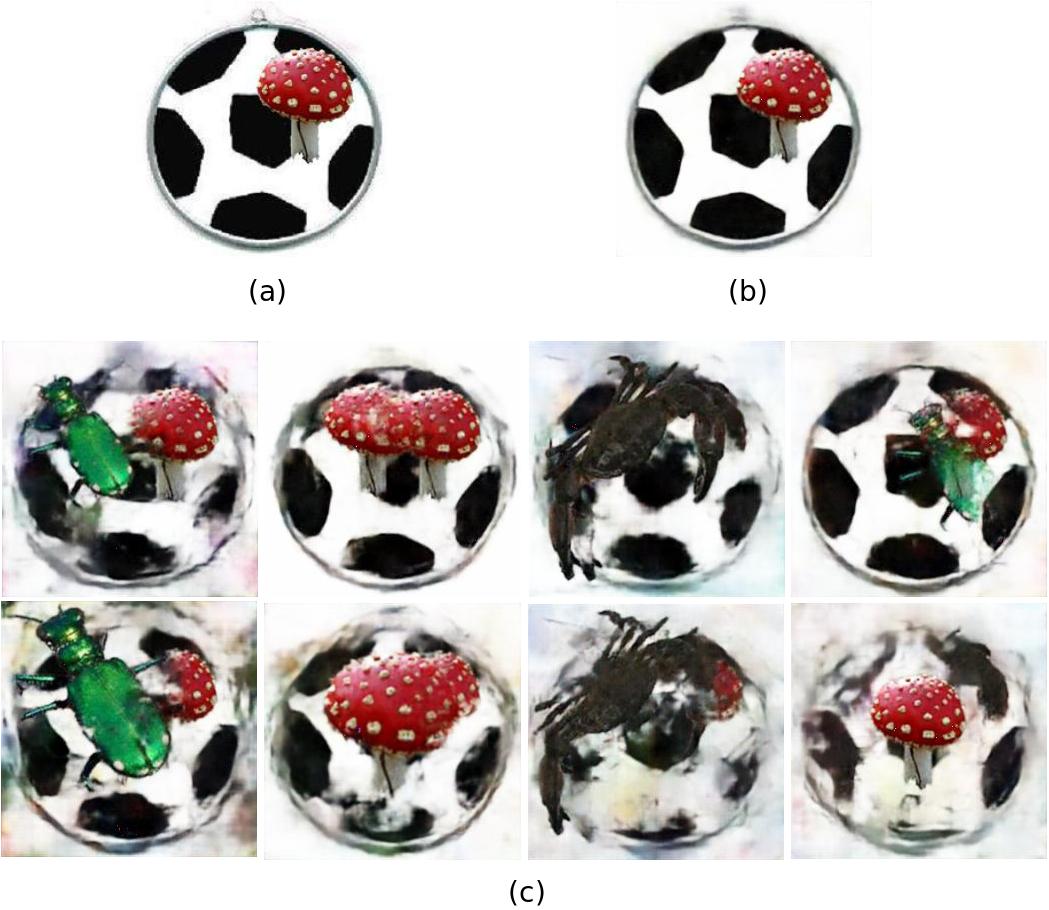}
	\caption{Visualization of the images reconstructed from the pseudo-DFVs. (a) the original synthetic occluded image of $\mathbf{v}_i$; (b) the image recovered by the DFV decoder from $\mathbf{v}_i$; (c) the example images reconstructed from pseudo-DFVs $\widetilde{\mathbf{v}}_{ik}$'s. The ResNet18\_C model in Section 4.1.2 is employed to extract the DFVs. The occlusion patterns shown in Fig.~5 are used to synthesize the occluded images. The clean and occluded image pairs outside the "soccer\_ball" class are adopted to generate the DVs $\mathbf{d}_k$'s and $\beta$ is set to $1$.}
	\label{sfig:occ_decoding}
\end{figure}

Fig.~S\ref{sfig:occ_decoding} shows the examples of the images reconstructed from the pseudo-DFVs generated from the DFV of a synthetic occluded image.
From Fig.~S\ref{sfig:occ_decoding}, we can observe that adding a DV to the DFV of an occluded image can result a pseudo-DFV corresponding to an occluded image with one or two occluders and all of these reconstructed images keep belonging to the "soccer\_ball" class.

\section{Additional Experiment - Oxford 102 flowers}
\begin{figure}[h]
	\centering
	\includegraphics[width=0.7\linewidth]{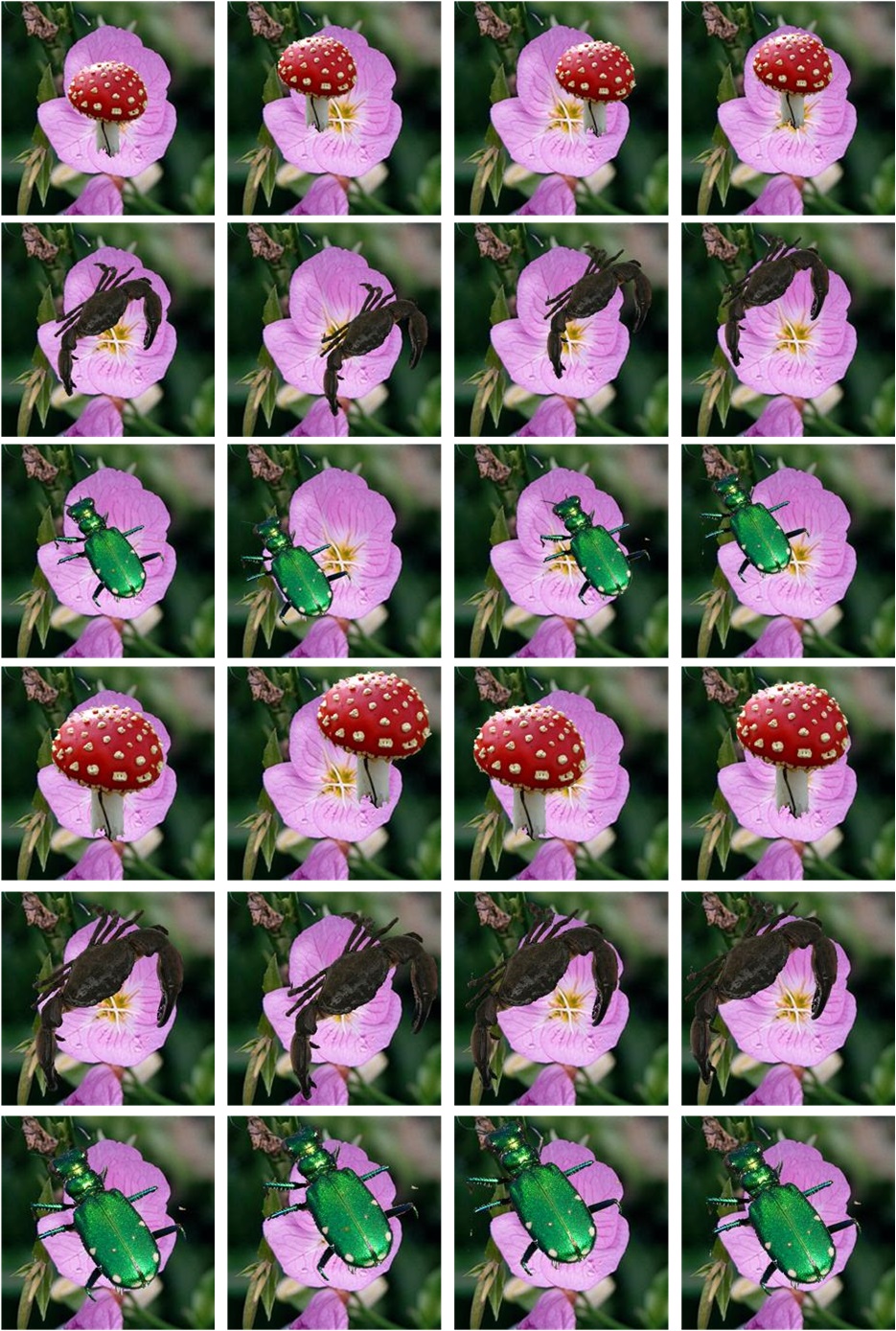}
	\caption{Examples of synthetic occluded test images for the experiments on the Oxford102 flower dataset. 
	}
	\label{fig:Oxford102flowerOccExamples}
	\vspace{10pt}
\end{figure}

In this section, we assess the proposed approach on a fine-grained dataset.
The Oxford 102 flower dataset is adopted for evaluation.
The Oxford 102 flower dataset consists of fine-grained image classes, 102 flower classes, with the number of images between $40$ and $258$ for each image class.
The settings for the experiment are similar to Section 4.1.2.
In particular, $\mathcal{I}_{op}$ is composed of $10$ image classes with $5$ images in each image class;
the set of clean training images $\mathcal{I}_{trn\_c}$ includes $102$ image classes with $20$ clean images each and $\mathcal{I}_{op\_c}$ is a subset of  $\mathcal{I}_{trn\_c}$;
the set of clean test images $\mathcal{I}_{tst\_c}$ contains $102$ image classes each a maximum of $50$ images excluding the clean images in $\mathcal{I}_{trn\_c}$;
the occlusion patterns used in the experiment are the same as those used in Section~4.1.
The examples of occluded images for each occlusion pattern are shown in Fig.~S\ref{fig:Oxford102flowerOccExamples}.

Four models following the naming protocol in Section 4.1.2, namely ResNet18\_F, ResNet18\_C, ResNet18\_C-Full and ResNet18\_F-Full, are evaluated for comparison. 
The classification results for each model and the improvements, ResNet18\_C-Full \wrt~ResNet18\_C and ResNet18\_F-Full \wrt~ResNet18\_F, are tabulated in Table \ref{tab:oxford102flowerResNet18_inclusive}.
The results demonstrate that the proposed approach significantly improves the classification of the occluded images for the fine-grained dataset. 
An interesting result is that the ResNet18\_C achieves better performance on the occluded images than that of the ResNet18\_F.
The reason might be that the occluded training images contaminated with occluders associated with the coarse-grained image classes are harmful to the model training on the fine-grained dataset. 

\begin{table}[]
	\renewcommand{\arraystretch}{1.3}
	\caption{
		Comparison of classification accuracy (\%) for the occlusion-inclusive training set on the Oxford 102 flower dataset
	}
	\centering
	\fontsize{8}{8}\selectfont
	\begin{tabular}{l|c|c}
		\hline
		Model         & Clean images            & \begin{tabular}[c]{@{}c@{}}Occluded images \\ Avg. over occlusion \end{tabular}         \\ \hline
		ResNet18\_C      & 92.11                   & 54.4                    \\ \hline
		ResNet18\_C-Full   & 92.16 & 66.75 ($\uparrow 12.35$)\\ \hline \hline
		ResNet18\_F    & 91.64                   & 51.07                   \\ \hline		
		ResNet18\_F-Full & 91.72 & 62.21 ($\uparrow 11.14$)\\ \hline
	\end{tabular}
	\label{tab:oxford102flowerResNet18_inclusive}
\end{table}

\section{Additional Experiment - CUB-200-2011} \label{sec:sec:cub2002011}
In this section, we compare the proposed DFV augmentation approach with the Attribute-Aware Attention Model ($\text{A}^3$M) \cite{han2018attribute} on Caltech-UCSD Birds-200-2011 (CUB-200-2011) dataset \cite{WahCUB_200_2011}.
The CUB-200-2011 dataset is a well-known benchmark for fine-grained classification algorithms. 
It contains 11,788 images of 200 classes. Each image is annotated with 15 part locations, 312 binary attributes, and 1 bounding box. 
The images cropped according to the object bounding box are used in the experiment.

\begin{figure}
	\centering
	\includegraphics[width=0.7\linewidth]{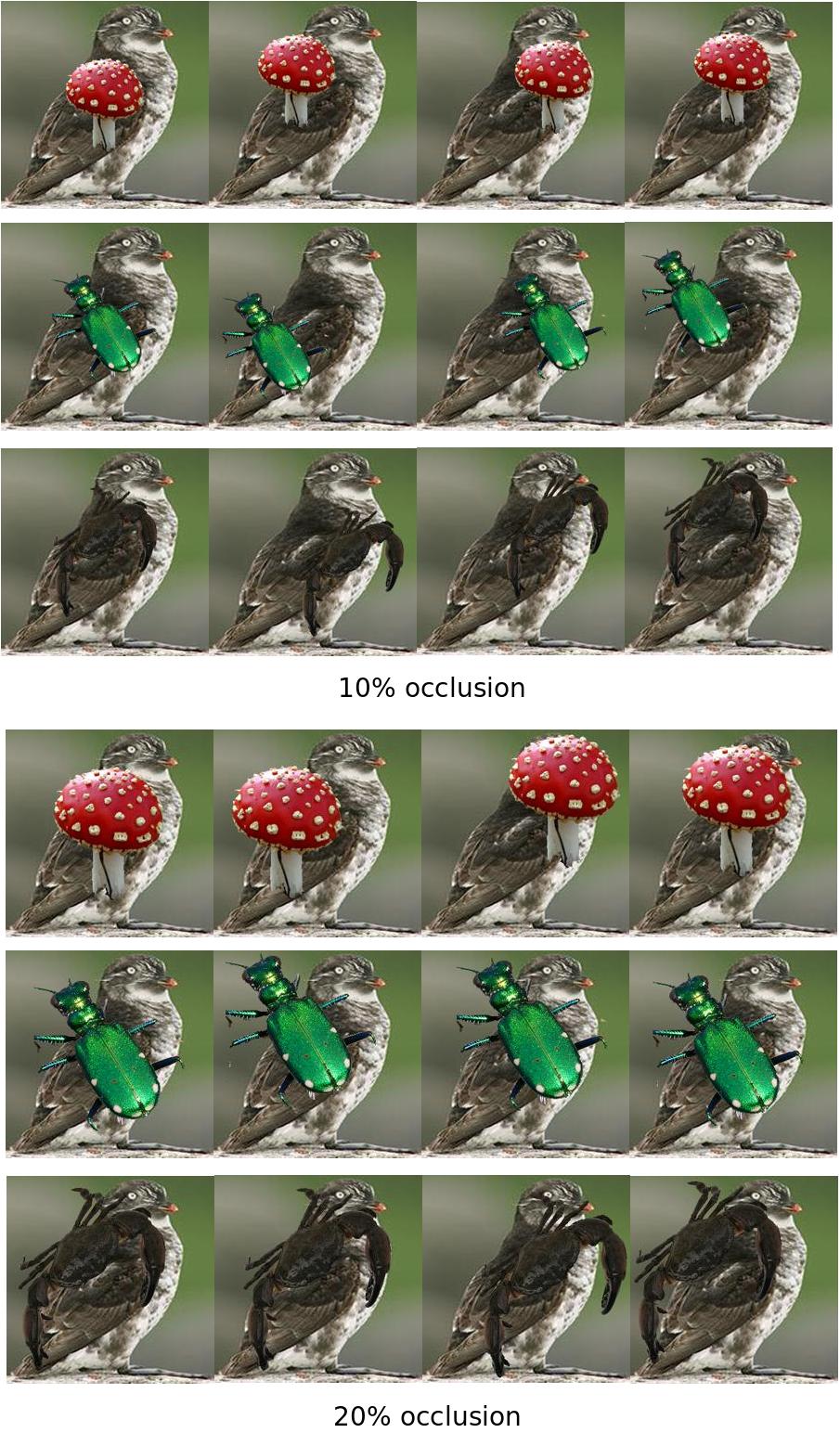}
	\caption{Example images for each occlusion pattern used for the evaluations on the CUB-200-2011 dataset}
	\label{fig:example_cub}
\end{figure}

The settings for the experiment are similar to Section~4.1.2.
The cropped images in the original training set and original test set are used as the set of clean training images $\mathcal{I}_{trn\_c}$ and the set of clean test images $\mathcal{I}_{tst\_c}$, respectively.
$\mathcal{I}_{op}$ is composed of 10 classes randomly drawn from the 200 classes and each with 5 clean images randomly drawn from $\mathcal{I}_{trn\_c}$.
The occluded images of $\mathcal{I}_{tst}$ and $\mathcal{I}_{op}$ are synthesized with 24 occlusion patterns shown in Fig.~5. 
The examples of occluded images that are resized to $224\times 224$ for each occlusion pattern are shown in Fig.~S\ref{fig:example_cub}.

The following four models are evaluated on $\mathcal{I}_{tst}$.
\begin{itemize}
	\item ResNet50\_C: ResNet50 network fine-tuned with the classical training approach on $\mathcal{I}_{trn\_c}$.
	\item ResNet50\_C-Full: ResNet50 network fine-tuned with the proposed approach on $\mathcal{I}_{trn\_c}$ and $\mathcal{I}_{op}$.
	\item $\text{A}^3$M\_s: $\text{A}^3$M with the input size of $224\times 224$ and trained on $\mathcal{I}_{trn\_c}$ by using the program provided by the author of \cite{han2018attribute} at \cite{iamhankai_a3m_codes}. 
	\item $\text{A}^3$M: $\text{A}^3$M with the input size of $448\times 448$ and trained on $\mathcal{I}_{trn\_c}$ by using the program provided by the author of \cite{han2018attribute} at \cite{iamhankai_a3m_codes} with default setting. 	
\end{itemize}
The ResNet50 network are adopted as the shared CNN for both $\text{A}^3$M\_s and $\text{A}^3$M.
The classification results are recorded in \tablename{ S\ref{tab:CUB_200_2011_occlusion_inclusive}}.

\begin{table*}
	\renewcommand{\arraystretch}{1.3}	
	\caption{Comparison of classification accuracy (\%) on the CUB-200-2011 dataset.}
	\centering
	\fontsize{8}{8}\selectfont
	\begin{tabular}{c|c|c|c|c|c}
		\hline
		Model          & Input size & Original images           & $20\%$ occlusion images   & $10\%$ occlusion images   & Avg. over occlusion       \\ \hline
		ResNet50\_C      & 224x224          & 81.86                     & 35.75                     & 61.09                     & 48.42                     \\ \hline
		ResNet50\_C-Full & 224x224          & 81.69 $(\downarrow 0.17)$ & 48.81 $(\uparrow 13.06)$  & 67.18 $(\uparrow 6.08)$   & 57.99 $(\uparrow 9.57)$   \\ \hline
		$\text{A}^3$M\_s        & 224x224          & 83.6 $(\uparrow 1.74)$    & 30.33 $(\downarrow 5.42)$ & 58.77 $(\downarrow 2.33)$ & 44.55 $(\downarrow 3.87)$ \\ \hline
		$\text{A}^3$M \cite{han2018attribute}       & 448x448          & 86.12 $(\uparrow 4.26)$   & 37.16 $(\uparrow 1.41)$   & 62.73 $(\uparrow 1.63)$   & 49.94 $(\uparrow 1.52)$   \\ \hline
	\end{tabular}
	\label{tab:CUB_200_2011_occlusion_inclusive}
\end{table*}

From \tablename{~S\ref{tab:CUB_200_2011_occlusion_inclusive}}, we can learn that the proposed approach significantly improve the classification accuracy on the occluded images (ResNet50\_C-Full vs. ResNet50\_C achieves $9.57\%$ increase) only with a very small degradation on the clean images (ResNet50\_C-Full vs. ResNet50\_C has $0.17\%$ decrease).

Compared to the ResNet50\_C model, both $\text{A}^3$M\_s and $\text{A}^3$M show considerable improvement on the clean images.
While for the occluded images, $\text{A}^3$M\_s, which has the same network input size as ResNet50\_C, shows a noticeable performance decrease.
Although $\text{A}^3$M vs. ResNet50\_C achieves a small performance increase on the occluded images, the input size of $\text{A}^3$M is larger than that of ResNet50\_C.
The $\text{A}^3$M model can focus on important or discriminative parts for different images by employing an attention mechanism.
However, if some of these parts are replaced by other objects, like the synthetic occluded images, where the occluder is another object, the $\text{A}^3$M model may still emphasize these parts leading to an increase of classification error.

{
\bibliographystyle{elsarticle-num}
\bibliography{arXiv_dfv_aug}
}
